%% file: main.tex
\PassOptionsToPackage{table,xcdraw}{xcolor}
\documentclass[sigconf]{acmart}
\usepackage{makecell}
\usepackage{subcaption}
\usepackage{multirow}
\usepackage{dsfont}

\acmYear{2023}\copyrightyear{2023}
\setcopyright{acmlicensed}
\acmConference[SenSys '23]{ACM Conference on Embedded Networked Sensor Systems}{November 12--17, 2023}{Istanbul, Turkiye}
\acmBooktitle{ACM Conference on Embedded Networked Sensor Systems (SenSys '23), November 12--17, 2023, Istanbul, Turkiye}
\acmPrice{15.00}
\acmDOI{10.1145/3625687.3625785}
\acmISBN{979-8-4007-0414-7/23/11}

\begin{document}

% \title{On Hallucinating Sensor Time-series Data for Counterfactual Training Conditions to Enhance Temporal Activity Classification} 
% \title{A Generative AI Framework for Time-series Data Generation in Unseen Conditions to Enhance Temporal Activity Classification}
\title{SudokuSens: Enhancing Deep Learning Robustness for IoT Sensing Applications using a Generative Approach}

% \author{
% Tianshi Wang, 
% Jinyang Li,
% Ruijie Wang,
% Denizhan Kara,
% Shengzhong Liu\texorpdfstring{$^{\dagger}$}{},
% Davis Wertheimer\texorpdfstring{$^{\S}$}{},
% Antoni Viros-i-Martin\texorpdfstring{$^{\S}$}{},
% Raghu Ganti\texorpdfstring{$^{\S}$}{},
% Mudhakar Srivatsa\texorpdfstring{$^{\S}$}{},
% Tarek Abdelzaher
% }
% \affiliation{University of Illinois at Urbana-Champaign \country{USA} } 
% \affiliation{$^{\dagger}$Shanghai Jiao Tong University \country{China}}
% \affiliation{$^{\S}$IBM T J Watson Research Center \country{USA}}
% \email{{tianshi3, jinyang7, ruijiew2, kara4}@illinois.ede, shengzhong@sjtu.edu.cn}
% \email{
% {Davis.Wertheimer, aviros}@ibm.com, {rganti, msrivats}@us.ibm.com, zaher@illinois.edu
% }
\author{Tianshi Wang}
\affiliation{University of Illinois Urbana-Champaign \country{USA}}
\email{tianshi3@illinois.edu}

\author{Jinyang Li}
\affiliation{University of Illinois Urbana-Champaign \country{USA}}
\email{jinyang7@illinois.edu}

\author{Ruijie Wang}
\affiliation{University of Illinois Urbana-Champaign \country{USA}}
\email{ruijiew2@illinois.edu}

\author{Denizhan Kara}
\affiliation{University of Illinois Urbana-Champaign \country{USA}}
\email{kara4@illinois.edu}

\author{Shengzhong Liu}
\affiliation{Shanghai Jiao Tong University\\ \country{China}}
\email{shengzhong@sjtu.edu.cn}

\author{Davis Wertheimer}
\affiliation{IBM T. J. Watson Research Center\\ \country{USA}}
\email{Davis.Wertheimer@ibm.com}

\author{Antoni Viros-i-Martin}
\affiliation{IBM T. J. Watson Research Center\\ \country{USA}}
\email{aviros@ibm.com}

\author{Raghu Ganti}
\affiliation{IBM T. J. Watson Research Center\\ \country{USA}}
\email{rganti@us.ibm.com}

\author{Mudhakar Srivatsa}
\affiliation{IBM T. J. Watson Research Center\\ \country{USA}}
\email{msrivats@us.ibm.com}

\author{Tarek Abdelzaher}
\affiliation{University of Illinois Urbana-Champaign \country{USA}}
\email{zaher@illinois.edu}

\authorsaddresses{}

\renewcommand{\shortauthors}{Tianshi et al.}





% \received{15 October 2023}
% \received[revised]{7 October March 2009}
% \received[accepted]{15 October 2023}
\input{content/abstract}

\maketitle
\input{content/introduction}

\input{content/motivation}

\input{content/methods}
\input{content/evaluation}
\input{content/related-work}

\input{content/conclusion}

\begin{acks}
This work was sponsored in part by ARL W911NF-17-2-0196, NSF CNS 20-38817, IBM (IIDAI), the Boeing Company, DARPA award HR001121C0165, DARPA award HR00112290105 and ACE (an SRC JUMP 2.0 Center). 
\end{acks}

\bibliographystyle{ACM-Reference-Format}
\bibliography{ref}

\end{document}

%% file: content/abstract.tex
\begin{abstract}
% partial condition sampling in IoT dataset. Reason: iot signals is a combination of object and env. A function of object and env. Image env change, cat not change. Vibration change. Much harder to collect a complete dataset for IoT. 
% Two sources of diversity: intrinsic attributes and temporal states
% Two approaches to handle: session-awared temporal (SA-TCLR) and Condition Interpolation.
This paper introduces {\em SudokuSens\/}, a generative framework for automated generation of training data in machine-learning-based Internet-of-Things (IoT) applications, such that the generated synthetic data mimic experimental configurations {\em not encountered\/} during actual sensor data collection. The framework improves the robustness of resulting deep learning models, and is intended for IoT applications where data collection is expensive. The work is motivated by the fact that IoT time-series data entangle the signatures of observed objects with the confounding intrinsic properties of the surrounding environment and the dynamic environmental disturbances experienced. To incorporate sufficient diversity into the IoT training data, one therefore needs to consider a combinatorial explosion of training cases that are multiplicative in the number of objects considered and the possible environmental conditions in which such objects may be encountered. 
Our framework substantially reduces these multiplicative training needs. To decouple object signatures from environmental conditions, we employ a Conditional Variational Autoencoder (CVAE) that allows us to reduce data collection needs from multiplicative to (nearly) linear, while synthetically generating (data for) the missing conditions. To obtain robustness with respect to dynamic disturbances, a session-aware temporal contrastive learning approach is taken. Integrating the aforementioned two approaches, {\em SudokuSens\/} significantly improves the robustness of deep learning for IoT applications. We explore the degree to which {\em SudokuSens\/} benefits downstream inference tasks in different data sets and discuss conditions under which the approach is particularly effective.

\end{abstract}

%% file: content/introduction.tex
\section{Introduction}
Modern machine learning has revolutionized sensing applications, but its success remains contingent on the availability of representative training data.
This paper is motivated by sensing applications where data collection remains expensive. For example, in a defense scenario, where sensors are trained to identify different types of vehicles from their seismic signatures, getting access to the right vehicles is logistically non-trivial. We say that such scenarios suffer from {\em data scarcity\/}~\cite{haresamudram2022assessing, kwon2020imutube}. In such scenarios, datasets for training and validation may not fully represent the complexity and diversity of real-world conditions. Machine learning models trained with such data may therefore exhibit catastrophic failures in novel conditions upon deployment~\cite{wang2022methodological, chen2021sensecollect}.

\begin{figure*}[h!]
\centering
\includegraphics[width=0.7\linewidth]{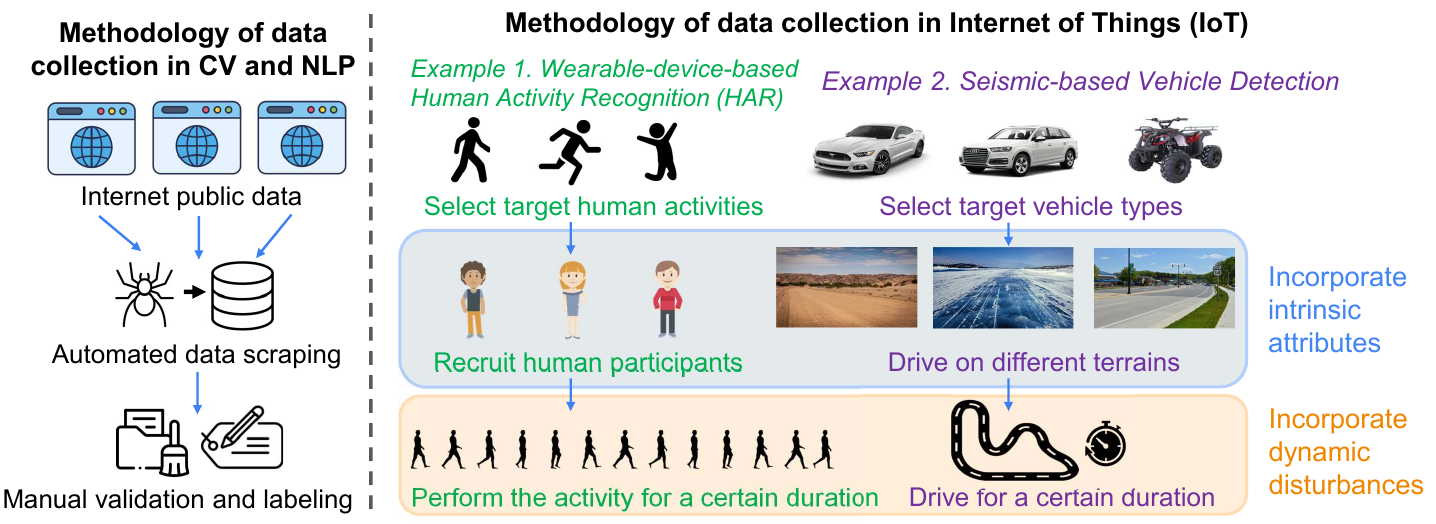}
\vspace{-0.05in}
\caption{The differences of dataset collection methodology between CV/NLP and IoT sensing applications}
\label{fig:data_collect}
\vspace{-0.1in}
\end{figure*}

The data scarcity problem is generally more pronounced in IoT applications compared to other fields such as computer vision (CV) and natural language processing (NLP). The specialized nature of some IoT sensors (compared to say, cameras or sources of text) often makes it harder to perform data collection at scale. 
% crowdsource data collection (the way one might advancement of the Internet leads to a daily surge of human generated data in forms of images, videos, text and speech\cite{data_never_sleeps}. Much of this data is naturally indexed by search engines, and even labeled by the publishers. Thus, researchers in these fields can benefit from large amounts of data and develop robust deep learning models that can achieve impressive performance in real-world applications. 
%
%However, the situation in IoT differs substantially. Even though vast amount of data are produced by IoT sensors each day, fully exploiting the data is challenging. IoT data is device-generated and lacks natural indexing on the Internet. The heterogeneity in device models, modailities, and setups poses further difficulties, lowering the compatibility between data sources. 
Moreover, due to the non-interpretable nature of IoT sensor signals, manually labeling data after the fact is difficult. 
%Even though companies (like Apple and Fitbit in IMU data about human activities\cite{perez2017privacy}) may possess large, clean datasets, they are often inaccessible to academic researchers due to privacy issues\cite{seliem2018towards}.

IoT applications involving specialized data thus need a different data collection methodology compared to CV and NLP. As shown in Figure~\ref{fig:data_collect}, the gathering of a dataset in CV and NLP usually involves scraping public data from the Internet in an automatic manner, followed by manual validation and labeling. A typical example is the construction of ImageNet~\cite{deng2009imagenet, yang2020towards}. 
%It first collects abundant images by query on search engines, and then relies on manual labor to filter the candidate images. 
In contrast, specialized IoT data collection often entails manually designing and conducting experiments for the express purpose of generating training data. For example,  vehicles might need to be driven at different distances from desired sensors in the intended types of terrain, for purposes of 
measuring their seismic signature in such terrains.
%To illustrate, for gathering a human activity recognition dataset, researchers have to recruit human participants, equip them with devices, and instruct them to perform various activities. For the seismic-sensor-based vehicle detection application, various vehicles need to be driven on different terrains. 
The time-consuming and costly nature of this process contributes to data scarcity, which impacts the quality of training for this category of IoT applications. 
%Typically, in an IoT application, a model is initially trained and validated on a dataset during what we term as the development phase. After its performance has been optimized, the model is deployed to real-world scenarios in the deployment phase.  Many researchers consequently have to develop IoT deep learning models under data scarcity. While many of these models may perform well in the development phase, they fail to generalize to real-world, diverse application scenarios in the deployment phase\cite{wang2022methodological}.

% \ruijie{I think we might need to define data scarcity issue a bit detailed in the Introduction, like "labeled samples cannot cover all diverse scenarios, or not all the data collected from each diverse scenario have high-quality labels."}

To address data scarcity -- specifically, the situation where the data available for developing a deep learning model are not sufficiently diverse, thereby failing to cover the full spectrum of conditions that the model may encounter during deployment -- 
our framework, {\em SudokuSens\/}, reduces the number of combinations of target and environment for which data must be physically collected. Data for the missing combinations are instead generated by our framework synthetically (much like solving a Sudoku puzzle), by combining clues from other {\em partially matching\/} conditions. Training can subsequently use both the physically collected and synthetically generated data, thereby mitigating the data scarcity challenge.

%We argue that addressing the gap between data diversity during model development and deployment is essential for practical deep learning models in IoT applications.  Therefore, we propose RobustIoT, a general framework aimed at bridging the gap between  datasets and real-world scenarios, enhancing the robustness of downstream deep learning models under data scarcity in IoT applications.

In this work, we adopt a broad definition of IoT applications, where the underlying sensing and data processing system is assumed to possess adequate capacity to conduct  ML-based inference tasks. This is in line with modern IoT applications, including activity recognition and target detection, where smartphones or (lower-end) edge servers can be enlisted to process sensing data, as contrasted with the emphasis on mote-class devices and low-end microcontrollers in the early days of sensor networks research. The analytic tasks performed by our applications were thus executed on a Raspberry Pi class device. 
The goal of these AI/inference tasks is to determine (in-situ) certain target attributes in the face of a variety of confounding conditions (static and dynamic) under which the underlying sensory observations are made. 
We call these confounding conditions {\em intrinsic attributes\/} and  {\em dynamic disturbances\/}, respectively, depending on whether they represent static discrete types of environment or dynamic conditions changing on a continuous scale.
For example, in the case of seismic sensing, an intrinsic attribute of the environment might be the type of terrain a vehicle drives on (e.g., asphalt, gravel, or dirt), whereas a dynamic disturbance might constitute the dynamic wind noise.

% a background activity, such as intermittent construction sounds on the side of the road.

%Intrinsic attributes refer to the constant characteristics of targets or their environments. Temporal states are the dynamic aspects of physical activities, changing over time. Ideally, a comprehensive IoT dataset should encompass adequate diversity from both of the two variance sources: data must be collected from various environments and objects over a sufficiently long duration. However, due to the difficulties in data collection, IoT datasets are typically gathered in a sub-optimal fashion. 
{\em SudokuSens\/} introduces two major novel components. First, to handle the diversity of intrinsic attributes, it uses a conditional variational autoencoder (CVAE)\cite{sohn2015learning, yan2016attribute2image} to augment the original dataset by interpolating the missing conditions. Thus, the IoT dataset need only sample a subset of all possible attribute combinations. For example, in a vehicle detection task, a training dataset may only cover the condition ``vehicle A on a city road", "vehicle A on a sandy road" and ``vehicle B on a city road". Conditional interpolation can then generate ``vehicle B on a sandy road". Conditional interpolation leverages the knowledge learned from the sampled conditions to synthesize the missing conditions. The purpose is to provide a more comprehensive augmented dataset. %This allows the downstream deep learning model to learn better underlying features rather than attribute-dependent features.

Second, to handle dynamic disturbances, {\em SudokuSens\/} employs a Session-Aware Temporal Contrastive Learning approach (SA-TCL). Since we assume that specialized IoT datasets are manually collected from specially designed experiments, we assume that the experiments are divided into sessions, where each session focuses on some specific physical activity in the presence of confounding, varying background states. For example, in a 30-minute ``vehicle A on a city road" data collection session, all samples correspond to vehicle type A. %Without sufficient data points, neural networks tend to memorize the existing samples instead of learning the underlying state-independent features. 
Accordingly, during training, SA-TCL rewards the encoder neural network for placing samples from the same session closer together in the latent space, while placing samples from different sessions further apart, thus encouraging the emerging latent representation to focus on session-label-specific features while ignoring confounding dynamic disturbances.

The preparation and use of {\em SudokuSens\/} involves (i) pre-training, (ii) fine-tuning, then (iii) deployment/testing.
During pre-training, {\em SudokuSens\/} first runs conditional data interpolation to augment the original dataset. Subsequently, it applies SA-TCL to the augmented dataset to train the SA-TCL encoder. Encoder training distills disturbance-resistant latent representations of the (augmented) input data. To fine-tune the framework to specific downstream inference tasks, the trained encoder of SA-TCL first maps incoming data to the (disturbance-resistant) latent space, then a downstream deep learning network is trained to decode from that space to the output of the task at hand. Finally, at deployment/test time, the SA-TCL encoder followed by the aforementioned decoder network jointly map from input data to output inference.

We evaluate the efficacy of SudokuSens at improving machine-learning outcomes for datasets that differ in their scarcity (i.e., their coverage of relevant conditions), the complexity of foreground activities performed, and the nature of static and dynamics confounding factors experienced. The analysis yields preliminary insights on deployment attributes correlated with the efficacy of the proposed approach. 
%We show that SudokuSens brings performance improvements in various IoT applications for different downstream classifiers.
Beyond existing dataset-based evaluation, we also conduct experimental studies, demonstrating how SudokuSens is integrated into real-world IoT sensing systems, where it contributes to improved robustness to unseen conditions. SudokuSens is shown to outperform the best baseline by 10.74\% to 26.87\% in accuracy under the conditions considered.

The rest of this paper is organized as follows. We empirically motivate the need for SudokuSens in Section~\ref{sec:motivation}. Section~\ref{sec:framework_design} describes its design. Section~\ref{sec:evaluation} presents evaluation results. Section~\ref{sec:related_work} covers related work. Finally, we conclude the paper in Section~\ref{sec:conclusion}.

%% file: content/motivation.tex
\section{Motivation: The Gap Between Lab and Wild}
\label{sec:motivation}

We present a motivating case study that underscores the performance gap arising between laboratory and real-world conditions due to data scarcity in IoT applications. The study comprises a vehicle detection application based on acoustic and seismic sensors. The goal is to detect the passage of a given vehicle type through the detection area by running a deep learning model on 2-second data intervals from both types of sensors. During data collection, both the seismic and acoustic sensors are on the ground, sensing the vibrations and sound generated when a vehicle is driven by.

\input{content/tables/data_collection_scenarios}

% Regarding to the data collection settings, as shown in Figure~\ref{fig:hardware}, a RaspberryShake 1D\cite{RaspberryShake}, which is a Raspberry Pi 4 connected with a seismic sensor, is adopted to collecting seismic signals. A ReSpeaker USB Mic Array\cite{mic_array} is used for collecting audio signals, which has 4 high performance digital microphones that support 360 degree audio pick-up. The microphone array is connected with the RaspberryShake via a USB cable. The Raspberry Pi collects seismic and acoustic signals at a sampling rate of 100 Hz and 8000 Hz respectively. The whole device is placed on a solid ground as depicted in Figure~\ref{fig:raspberryshake}. This sensor device streams the data to a nearby laptop via 2.4 GHz Wi-Fi signal. The Lenovo ThinkPad T430 laptop (shown in Figure~\ref{fig:edge}), as the edge device, runs vehicle detection model inference on 2-second data chunks on the received sensor data in real-time. 

%To gather the dataset, we follow the commonly employed manual collection methodology in IoT. 
We perform the data collection in seven scenarios, described in Table~\ref{tab:scenarios}. Each scenario is a unique combination of location and confounding factors that introduce variability to challenge the classification task. %We show the locations in the seven scenarios in Figure~\ref{fig:scenarios}.
A total of eight vehicle types were utilized. Constrained by both financial and temporal budgets, our data collection could only partially cover the full range of possible vehicle-environment combinations, as displayed in Table~\ref{tab:sampled_conditions}. In each data collection session, the vehicle is arbitrarily driven at a speed between 5 miles per hour to 25 miles per hour within a radius of 300 feet around the sensors. In total, we collected 17 sessions of data for model development.

\input{content/tables/sample_conditions}

\begin{figure}[h]
\centering
\includegraphics[width=0.85\linewidth]{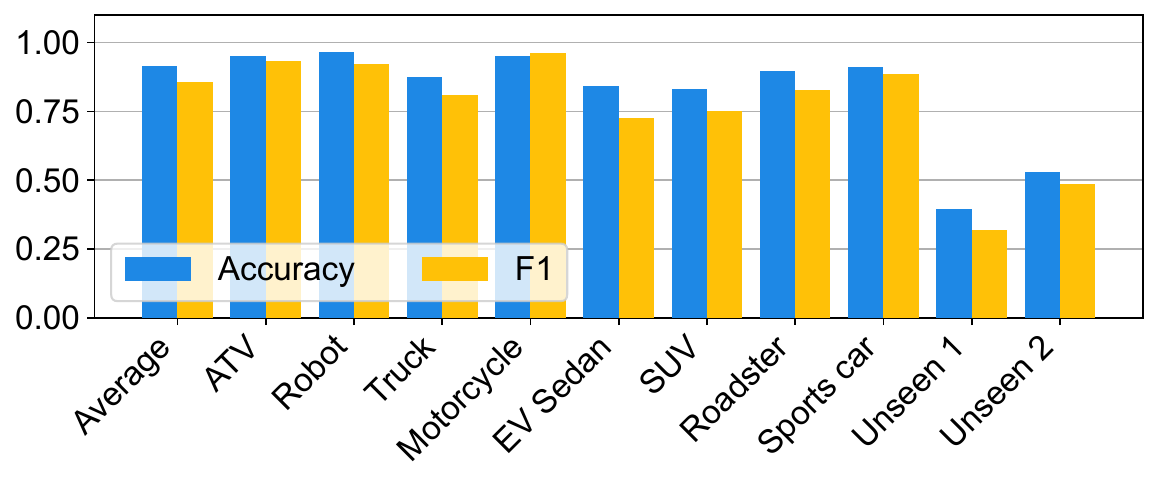}
\vspace{-0.2in}
\caption{A case study to motivate SudokuSens. The average accuracy and the accuracy for each vehicle type during the development phase are presented.}
\label{fig:case_study}
\end{figure}

The data set offers
incomplete coverage of all possible vehicle-environment combinations. 
%This partial coverage reduces training accuracy. Second, the 20-minute duration for each data collection session fails to capture the full extent of dynamic runtime conditions possible, including the speed and the position of the car, as well as the environmental noise. We develop a deep learning model that is optimized based on the development dataset. 
We split it into a training set, a validation set, and a testing set in the proportions 80\%, 10\%, and 10\%, respectively, by splitting the sensor traces from each condition into three contiguous partitions whose lengths are of the above proportions to be included respectively in the corresponding sets. Thus, (some part of) each condition is represented in each of the training, validation, and testing data. For the classifier, we implemented a deep learning model based on DeepSense~\cite{yao2017deepsense}, a supervised neural network designed for time-frequency learning from IoT signals. We then optimized this network by doing a manual neural architecture search over the number of layers (including CNNs and RNNs), the length of the feature dimensions, dropout rates, and learning rates. Additionally, we adjusted the parameters of the short-time Fourier transform (STFT) applied to the input data. The accuracy and F1 score of the developed classifier (applied to test data) for each of the eight targets, as well as their average, is shown in Figure~\ref{fig:case_study}.

As can be seen from Figure~\ref{fig:case_study}, when test data are drawn from the same environmental conditions as training and validation, the performance is acceptable. To evaluate the robustness of the model, we then collect data on two {\em additional scenarios\/}, featuring the same targets in new conditions (intrinsic attributes and dynamic disturbances not used for training/validation). %Data for each of the other scenarios is partitioned as usual between training, validation, and testing. 
In Figure~\ref{fig:case_study}, we call them {\em Unseen~1\/} and {\em Unseen~2\/}.
The goal is to evaluate how well the trained model generalizes to previously unseen conditions (i.e., conditions not explicitly trained with). 

\subsection{Unseen Condition 1: Different Intrinsic Attributes}
\label{subsec:case_study1}
For {\em unseen condition~1\/}, we drive (during testing) a Ford Mustang in scenario~F. It is a previously unseen combination of vehicle, background traffic, and terrain type. 
As shown in Figure~\ref{fig:case_study}, the model exhibits poor performance in this case, compared to testing in scenarios used in training. Even though the new condition involves both (i) a vehicle type that appeared in training (under other environmental conditions) and (ii) environmental conditions that appeared in training (for other vehicles), the training tends to overfit to the exact seen target/environment combinations, failing to disentangle the influences of target and environment. One might potentially blame the training, but fundamentally, the observed overfitting is a result of data scarcity. Given a modern neural network and limited training data, there are enough neural network parameters to ``memorize" the training data, thereby failing to generalize.
This outcome highlights the limitations of training a deep learning model on scarce data, as the trained model may then lack the robustness to generalize well.

\subsection{Unseen Condition 2: New Dynamic Disturbances}
\label{subsec:case_study2}
For {\em unseen condition 2\/}, we drive the Ford Mustang in the location of scenario B but on another day with a high wind noise (which departs from conditions of scenario B). The strength and direction of the wind were highly dynamic during the experiment. As shown in Figure~\ref{fig:case_study}, there is a significant drop in accuracy under the new condition, compared to the average accuracy for the conditions trained with. 

The observations in the above two cases motivate us to enhance the robustness of deep learning models trained with scarce data. Below, we present the general framework of SudokuSens.

%% file: content/tables/data_collection_scenarios.tex
\begin{table}[h]
\caption{Data Collection Scenarios.}
\vspace{-0.15in}
\label{tab:scenarios}
\scriptsize
\resizebox{1.0\linewidth}{!}{
\fontsize{8.5}{11}\selectfont
\begin{tabular}{r|lllll}
\hline
& \textbf{Location}     & \textbf{Description}    & \textbf{Terrain} & \textbf{Traffic} & \textbf{Wind}\\
\hline
A                 & Parking structure     & Roof top parking        & Concrete         & Low                    & High \\
B                 & College parking lot   & Large outdoor parking   & Concrete         & Low                    & Medium\\
C                 & Stadium parking lot 1 & Small outdoor parking   & Concrete         & Medium                 & Low \\
D                 & City parking lot      & Small outdoor parking   & Concrete         & Medium                 & Medium \\

E                 & State park            & Clearing in wooded area & Gravel           & No                     & Medium \\
F                 & Stadium parking lot 2 & Large outdoor parking   & Gravel           & Medium                 & Low \\
G                 & Undeveloped area      & Overflow parking        & Gravel           & High                   & Low \\
\hline
\end{tabular}}
\end{table}

%% file: content/tables/sample_conditions.tex
\begin{table}[h]
\small
\caption{Experimental conditions in the vehicle detection dataset. 
Each cell shows the minutes of data collected under that condition.  Blank cells are conditions not sampled.}
\vspace{-0.1in}
\label{tab:sampled_conditions}
\resizebox{1.0\linewidth}{!}{
\fontsize{8.5}{11}\selectfont
\begin{tabular}{rllllllll}
\hline

& \textbf{Type} & \textbf{A} & \textbf{B} & \textbf{C} & \textbf{D} & \textbf{E} & \textbf{F} & \textbf{G}
\\
\hline

% Name                            Type           A   B  C  D  E  F  G
\textbf{Polaris}                & ATV           &  &  &  &  &20&  & \\
\textbf{Warthog}                & Robot         &  &  &  &  &20&  & \\
\textbf{Chevrolet Silverado}    & Pickup Truck  &  &60&  &  &15&  & \\
\textbf{Ducati Scrambler}       & Motorcycle    &  &60&  &  &  &30& \\
\textbf{Tesla Model 3}          & EV Sedan      &  &70&  &  &  &  & \\
\textbf{Nissan Rogue}           & SUV           &  &  &60&  &  &20& \\
\textbf{Mazda MX-5}             & Roadster      &30&  &30&  &  &30& \\
\textbf{Ford Mustang}           & Sports Car    &30&60&30&20&  &  &20\\
\hline
\end{tabular}}
\end{table}

%% file: content/methods.tex
\section{Framework Design}
\label{sec:framework_design}
In this section, we describe the design of SudokuSens that addresses the robustness challenge. We first give an overview of the whole framework then describe each of its two key components in detail.

\begin{figure*}[h!]
\centering
\includegraphics[width=\linewidth]{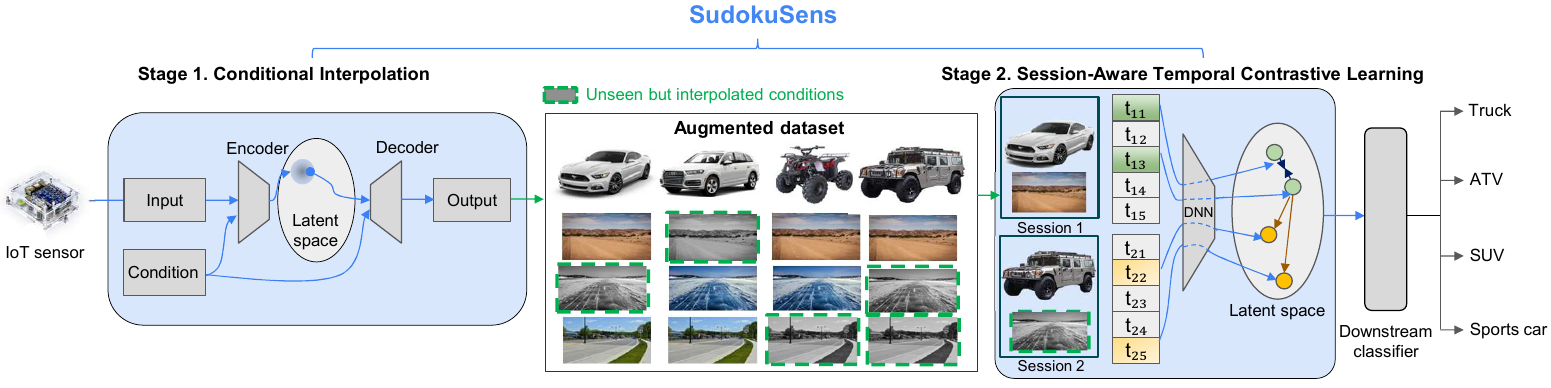}
\caption{SudokuSens. The name highlights the design of conditional interpolation; it generates data for missing intrinsic attribute combinations much like solving a Sudoku puzzle.}
\label{fig:SudokuSens}
\end{figure*}

\subsection{SudokuSens Overview}
As depicted in Figure~\ref{fig:SudokuSens}, in the offline pre-training stage, a training dataset is first used to train a CVAE model to disentangle the influences caused by various intrinsic environmental attributes. Later, the trained CVAE is used to synthetically generate data for conditions missing from the original dataset. The synthetic samples together with the original samples form the augmented dataset. Next, SA-TCL takes the augmented dataset as input and applies contrastive learning. It pulls samples from the same sessions closer together in the latent space and repels samples from different sessions apart. The trained encoder in SA-TCL is then frozen, and prepended to a downstream classifier as an initial feature extractor to mitigate the effects of dynamic disturbances. The downstream classifier is then trained using the augmented dataset. Finally, during deployment/testing, each input data sample is passed through the SA-TCL encoder and then fed into the downstream classifier to accomplish the run-time inference task. 

\subsection{Conditional Interpolation}

\begin{figure}[h]
\centering
\includegraphics[width=0.8\linewidth]{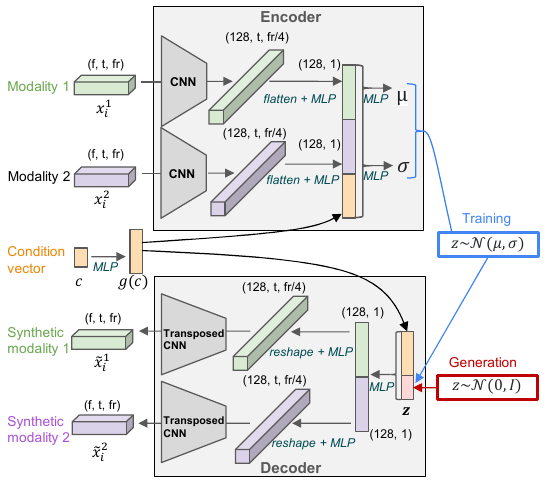}
\caption{The CVAE architecture for conditional interpolation. A 2-modality input is taken as an example to illustrate the design.}
\label{fig:condinterp}
\end{figure}

Let us define $\mathcal{D}$ as the original dataset containing signals $\mathbf{x}$ with their associated conditions $\mathbf{c}$, such that $\mathcal{D} = {(\mathbf{x}_i, \mathbf{c}_i)}_{i=1}^{N}$, where $N$ is the total number of samples in the dataset. Each condition $c_i$ is a vector that represents a specific combination of attribute values, where each attribute value is one specific value from an attribute in the set $\mathcal{A}$. For instance, in the vehicle detection application, if $\mathcal{A} = \{\text{{vehicle type}, \text{{terrain type}}}\}$, then a condition $c_i$ could be $\left(\text{{sedan}}, \text{{desert}}\right)$, representing a specific vehicle type and a specific terrain type. Furthermore, we denote the set of all seen conditions in the dataset as $\mathcal{C}_{\text{{seen}}}$, and the set of missing conditions as $\mathcal{C}_{\text{{unseen}}}$. Each $\mathbf{c}_i \in \mathcal{C}_{\text{{unseen}}}$ represents a combination of attribute values that do not exist in $\mathcal{C}_{\text{{seen}}}$, although each individual attribute value has appeared in some combination within $\mathcal{D}$.

As shown in Figure~\ref{fig:condinterp}, the CVAE takes multi-modality signals as inputs, so $\mathbf{x}_i = \{\mathbf{x}^j_i\}_{j=1}^{M}$, where each $\mathbf{x}^j_i$ represents the signal from the $j$-th modality and $M$ is the total number of modalities. We preprocess the time-series raw data by short-time Fourier transform (STFT), which effectively exposes the patterns in the frequency domain~\cite{yao2019stfnets}. Thus, the input from each modality is in the shape of $\mathit{feature} \times \mathit{time} \times \mathit{frequency}$, annotated as $f$, $t$, $\mathit{fr}$ respectively (we annotate the inputs to the neural network using the Channels First format.). During training, signals $\mathbf{x}$ are fed into the encoder. Signals from each modality first pass through multiple convolutional layers that reduce the frequency dimension and capture the essential features in a lower-dimensional space.
After that, feature maps from each modality are flattened and concatenated alongside the conditional feature $g(\mathbf{c})$.
Conditional feature $g(\mathbf{c})$ is derived from the conditional vector $\mathbf{c}$ passing through a multi-layer perceptron (MLP), where $g(\cdot)$ represents the function of the MLP.
In our implementation, we one-hot embed each attribute value, and concatenate them into the condition vector $\mathbf{c}$.
Next, the concatenated vector passes through two separate MLPs to generate the mean $\mu$ and standard deviation $\sigma$ of a Gaussian distribution respectively.
Finally, a latent representation $\mathbf{z}$ is randomly sampled from $\mathbf{z} \sim \mathcal{N}(\mu, \sigma)$. 

The objective of the decoder is to reconstruct the original signals given the latent representation $\mathbf{z}$ alongside the conditional feature $g(\mathbf{c})$.
The decoder maps the latent representation $\mathbf{z}$ to a higher dimensional tensor by multiple MLPs and processes the feature map of each modality by several transposed convolutional layers.
The reconstructed signals for each modality have the same shape as the inputs.
%
% The optimization target of the CVAE:
Following common practice, we utilize evidence lower-bound (ELBO)~\cite{sohn2015learning, DyDiff-VAE, infovgae, NEURIPS2022_6b295b08} to optimize CVAE model parameters:
\begin{equation}
\small
\begin{split}
\mathcal{L}(\theta, \phi; \mathbf{x}, \mathbf{c}) &= -\mathbb{E}_{q_{\phi}(\mathbf{z}|\mathbf{x}, \mathbf{c})}[\log p_{\theta}(\mathbf{x}|\mathbf{z}, \mathbf{c})] \\
&\qquad+ KL(q_{\phi}(\mathbf{z}|\mathbf{x}, \mathbf{c}) || p(\mathbf{z})).
\end{split}
\label{equation:cvae_train}
\end{equation}
\noindent
The first term of the loss function represents the reconstruction loss. It ensures that the CVAE learns to accurately reconstruct the input data. Here, $q_{\phi}(\mathbf{z}|\mathbf{x}, \mathbf{c})$ is the probability distribution that the encoder uses to map the input data $\mathbf{x}$ and the conditional vector $\mathbf{c}$ to the latent space, and $p_{\theta}(\mathbf{x}|\mathbf{z}, \mathbf{c})$ is the probability distribution that the decoder uses to map from the latent space back to the data space. The second term is the Kullback-Leibler (KL) divergence between the encoder's distribution $q_{\phi}(\mathbf{z}|\mathbf{x}, \mathbf{c})$ and a prior distribution $p(\mathbf{z})$, which is chosen to be a standard normal distribution. The KL divergence measures how much the encoder's distribution over the latent space deviates from the prior distribution. This term acts as a regularization that encourages the distribution of latent variables to be close to the prior distribution, preventing the model from overfitting.

Through training, the CVAE learns the influences of different intrinsic attributes in $\mathcal{A}$.
Thus, after training, conditional interpolation is applied to $\mathcal{C}_{unseen}$. During the interpolation, latent representation $\mathbf{z}$ is directly sampled from a standard normal distribution $\mathbf{z} \sim \mathcal{N}(0, I)$ then passed through the decoder together with a $\mathbf{c} \in \mathcal{C}_{unseen}$ to generate the synthetic sample $\hat{\mathbf{x}}$.

\begin{equation}
\hat{\mathbf{x}} = p_{\theta}(\mathbf{z}, \mathbf{c}).\label{equation:cvae_generation}
\end{equation}

To incorporate sufficient variances, we interpolate a given $\mathbf{c}\in \mathcal{C}_{unseen}$ for $T$ times. The synthetic samples alongside the original dataset forms the augmented dataset $\mathcal{D}_{aug}$.

\subsection{Session-Aware Temporal Contrastive Learning}

Session-aware temporal contrastive learning (SA-TCL) leverages the fact that IoT datasets are often organized by data collection sessions.
Samples within the same session reflect the same physical phenomenon in the presence of various dynamic disturbances.
% However, due to insufficient data sampling, a limited range of variances in temporal states are captured in the dataset.
It is advantageous to guide the neural network to learn disturbance-independent features to avoid overfitting. 

\begin{figure}[h]
% \vspace{-0.3in}
\centering
\includegraphics[width=0.8\linewidth]{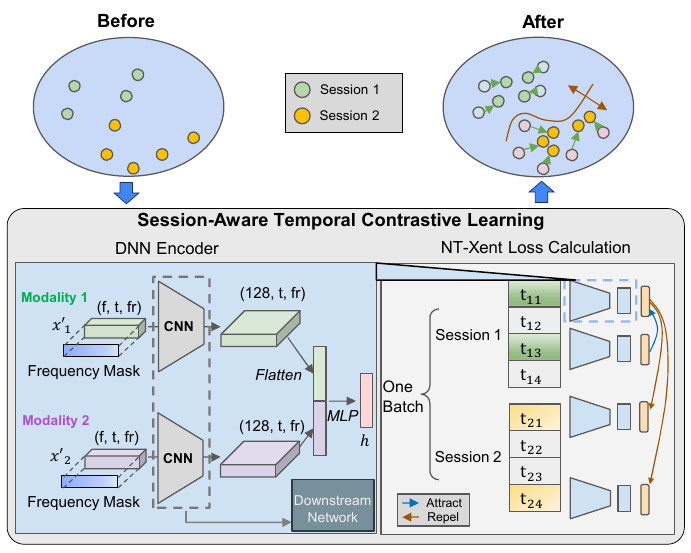}
\caption{Session-aware temporal contrastive learning.}
\label{fig:SA-TCL}
\vspace{-0.15in}
\end{figure}

As shown in Figure~\ref{fig:SA-TCL}, the encoder takes samples from the conditional interpolated dataset $\mathcal{D}_{aug}$ as inputs.
We annotate each sample from the $\mathcal{D}_{aug}$ as $\mathbf{x}^\prime_{i}$.
As a multi-modality input, $\mathbf{x}^{\prime j}_{i}$ represents the signal of the $j$-th modality from sample $i$. An STFT is computed to convert the time-series data into a spectrogram. During contrastive learning, the different frequency components of the spectrogram could be weighted differently based on the information density. For example, 
in the applications we consider (such as vehicle detection and human activity recognition), the lower-frequency part of the STFT spectrogram contains more target-related information than the high-frequency part.
% During contrastive learning, distance at different part of the spectrograms should be weighted based on the information density.

Thus, we apply a frequency mask $M$ to each $\mathbf{x}^{\prime j}_{i}$ to put different weights along the frequency axis of the original spectrogram.
In the frequency mask, the values closer to the lower frequency are initialized closer to 1, while the values closer to the higher frequency are initialized closer to 0. For values in between, they are initialized using a descending logarithmic scale, transitioning smoothly from values near 1 to values approaching 0.
% By multiplying the frequency mask to the original spectrograms, the distance between samples in the lower frequency part is highlighted and the higher frequency part is diminished.
By multiplying the frequency mask by the original spectrogram, the contribution of low-frequency components in computing the distance between samples is emphasized, while the contribution of high-frequency components is weakened.
We define the masked sample $\tilde{\mathbf{x}}^j_i$ as:
\begin{equation}
\tilde{\mathbf{x}}^j_i = M \cdot {\mathbf{x}}^{\prime j}_i.
\end{equation}

As different tasks may have different information density distributions, the values in $M$ are set as learnable parameters, so that the model has the flexibility to adjust this mask and learn the best way to focus on different parts of the spectrogram. 

After the frequency mask is applied, the signals from each modality pass through multiple convolutional layers that keep the original shape of the spectrogram but increase the feature dimension to expose features in a higher dimensional space. Then, the output tensors from each modality are flattened and concatenated into one vector $\mathbf{z}$. This vector further goes through an MLP and produces the latent representation for calculating the contrastive loss.

The computation of contrastive loss is based on the Normalized Temperature-scaled Cross Entropy loss (NT-Xent loss)~\cite{sohn2016improved, chen2020simple}. Given a batch size $B$, within each training iteration, $B$ sessions are randomly selected. Within each session, two different samples are randomly chosen, making $2B$ samples in total for each batch. When calculating the contrastive loss, for each sample $i$, the sample $j$ from the same session is treated as the positive sample, while the remaining $2B-2$ samples are considered negative. The latent representations of all the samples are calculated through the encoder and annotated as $h_i$. The NT-Xent contrastive loss of SA-TCL  for a positive pair $(i, j)$ can be defined as:

\begin{equation}
\ell(i, j) = -\log \frac{\exp\left(\text{sim}(h_i, h_{j})/\tau\right)}{\sum^{2N}_{k=1}\mathds{1}_{\left[k\neq i\right]}\exp\left(\text{sim}(h_i, h_{k})/\tau\right)}.
\end{equation}
And the total loss of a batch is the mean of the losses for all the positive pairs:
\begin{equation}
\mathcal{L} = \frac{1}{2N}\sum^{N}_{k=1}\left[ \ell\left( 2k-1, 2k\right) + \ell\left( 2k, 2k-1 \right) \right].
\end{equation}

%% file: content/evaluation.tex
\section{evaluation}
\label{sec:evaluation}
%In this section, we present the evaluation results on both datasets and case studies. 
We implement SudokuSens in Python based on PyTorch 1.11~\cite{pytorch}. The {\em training\/} of the framework is done on a server equipped with an Intel i9-9960X @3.10GHz CPU and 4 NVIDIA GeForce RTX 2080 Ti GPUs. At {\em deployment/inference\/} time, SudokuSens is run on a Raspberry Pi 4 (including the encoder and the downstream classifier). 

Below, we first introduce the experimental setup, including the datasets, the downstream neural networks, and the baselines.
We then present and discuss the overall dataset-based evaluation results.
Subsequently, we revisit the motivating experiments mentioned in Section~\ref{sec:motivation},
describe our implemented vehicle detection system in more detail, and show how SudokuSens enhances model robustness for this application during actual deployment.
Finally, we conduct ablation studies to investigate the factors affecting the degree of improvement brought by SudokuSens and present
profiling results to evaluate its run-time efficiency on the Raspberry Pi.

\subsection{Datasets}
\label{sec:dataset}
In the first of the evaluation, we conduct experiments on three typical IoT datasets. 
In each case, we map the data set into a matrix we henceforth suggestively call, the {\em Sudoku matrix\/}. One dimension of this matrix represents different classes that the classifier in question needs to distinguish (e.g., target types or human activities). The other dimension represents a set of different discrete conditions under which these classes might be observed. For evaluation purposes, we mark some of these target/condition combinations (i.e., matrix cells) as {\em seen\/} and others as {\em unseen\/}. We then divide each dataset into three subsets: training, validation, and testing. The training and validation sets draw {\em only\/} from cells marked seen, $\mathcal{C}_{seen}$, whereas the testing set comprises {\em only\/} the cells marked unseen, $\mathcal{C}_{unseen}$. The percent of the Sudoku matrix cells marked as {\em seen\/} is thereafter referred to as the {\em percent coverage\/} of the training/validation data. We did not explicitly control for dynamic disturbances, although natural noise in the data caused variations among different traces, even under the same conditions. Thus, such disturbances were naturally present. The evaluation is performed by performing classification of known targets in {\em unseen\/} conditions and computing the average accuracy and macro F1 score. Below, we describe the data sets in more detail.
% To manage the unseen conditions, we randomly masking certain conditions from the training/validation sets. We also ensure every attribute value is represented in the training set at least once, enabling information propagation from seen to unseen conditions. For each evaluation, we repeat the random masking 10 times and calculate the metrics by averaging.  

\begin{itemize}
\item
\textbf{Seismic- and acoustic-based vehicle detection}.
% \subsubsection{Seismic and audio-based vehicle detection}
The vehicle detection dataset we use is the Acoustic-seismic Classification Identification Data Set (ACIDS)~\cite{bennett2018cloud}.
ACIDS uses a seismic and an acoustic sensor, both at a 1024 Hz sampling rate, to record vehicles moving towards and away from the sensing zone. On average, the dataset contains $45$ minutes of recordings (per sensor) per target class, segmented into two-second chunks with one-second overlap. It encompasses 135 sessions from 9 vehicle types across 3 terrains: {\em arctic\/} (snow/ice), {\em normal\/} (city roads), and {\em desert\/} (sand). The purpose, in this case, is to do vehicle classification. Confounding conditions represent the three different terrain types. Clearly, the terrain type (e.g., asphalt versus snow) affects both the sound and vibration features, making it a proper confounding attribute.

\item \textbf{Wearable-device-based human activity recognition}.
% \subsubsection{Wearable-device-based human activity recognition}
We use the RealWorld-HAR dataset~\cite{sztyler2016body} for wearable-device-based human activity recognition, which captures data from 6 modalities, including accelerometer, GPS, gyroscope, light, magnetic field, and sound level. Sensors are placed on 7 body positions. On average, the dataset contains approximately one hour of recordings (per sensor) per activity class. It features 5 activities including walking, running, stairs up/down, and jumping, recorded from 15 human subjects. Each session lasts around 10 minutes (except for jumping, which lasts approximately 1.7 minutes). The purpose, in this case, is to perform human activity recognition. The confounding variable is the person performing the activity.

\item \textbf{Wireless-sensor-based human activity recognition}.
% \subsubsection{Wireless-sensor-based human activity recognition}
We use the wireless-sensor-based human activity recognition dataset from \cite{baha2020dataset}.
It records Wi-Fi signal variations caused by indoor human activities. On average, the data set contains approximately 15 minutes of recordings per activity class. It features 12 activities, performed by 30 human subjects across 3 environments.
Each session involves 20 repetitions of activities. The purpose, as before, is to perform human activity recognition. The confounding variable is, again, the person.
\end{itemize}

\subsection{Downstream Classifiers}
As a general feature extraction framework, SudokuSens can support different downstream classifiers. We select 3 typical neural network architectures as the downstream classifiers.

\begin{itemize}
\item    
\textbf{Shallow neural network}.
% \subsubsection{Shallow neural network}
The shallow neural network (denoted as \textit{shallow}) concatenates all the features from each modality together, and flattens the concatenated features into a 1D vector.
The network has a fully-connected hidden layer, followed by a ReLU activation function.
After that, a fully-connected layer and a softmax function are used for the final classification.
We choose this architecture to offer a baseline to compare against more complex architectures. 
\item
\textbf{DeepSense}.
% \subsubsection{DeepSense}
DeepSense~\cite{yao2017deepsense} is a neural network designed for IoT applications.
It extracts features from the input in a hierarchical way.
It first uses 3 convolutional layers to extract the modality-level features, then stacks the features maps and averages across modalities to finish modality fusion.
Next, if multiple sensor node locations exist (like in RealWorld-HAR), DeepSense follows the same pattern to do location fusion.
The fused features then pass through 1 Gated Recurrent Unit (GRU) layer to better extract the temporal related features.
Finally, a linear layer followed by a softmax acts as the classification head and outputs the inference result. 
We choose DeepSense to evaluate how SudokuSens can improve on a more complex neural network based on conventional network building blocks.
\item
\textbf{Transformer}.
% \subsubsection{Transformer}
In the Transformer network, the input spectrogram of each modality is first reshaped from $(f, t, \mathit{fr})$ to $(\mathit{fr} \times f, t)$ (Channels First format), and passed through a standard Transformer encoder layer\cite{vaswani2017attention}, which includes a self-attention layer and two linear layers.
Here, time dimension $t$ is the sequence length, and the product of frequency and feature dimension $\mathit{fr} \times f$ is taken as the embedding dimension.
Subsequently, multimodal features are fused via concatenation and linear layer processing.
If multiple locations exist, the network follows the same manner to further extract and fuse the location-level features.
Finally, a linear layer followed by a softmax acts as the classification head.
As one of the current mainstream and most successful architectures, we choose transformer to evaluate whether our method can further improve the performance of the state-of-the-art architecture.
\end{itemize}

\begin{figure*}[h!]
     \centering
     % Shallow neural network as downstream

     \begin{subfigure}[t]{0.5\textwidth}
         \centering 
         \includegraphics[width=\textwidth]{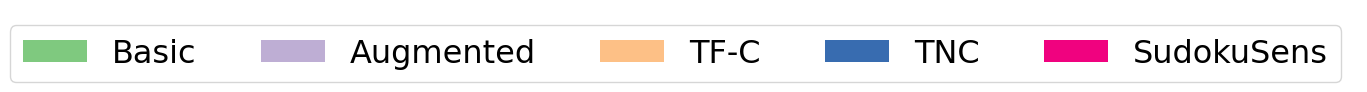}
    \end{subfigure}
    
     \begin{subfigure}[t]{0.31\textwidth}
         \centering 
         \includegraphics[width=\textwidth]{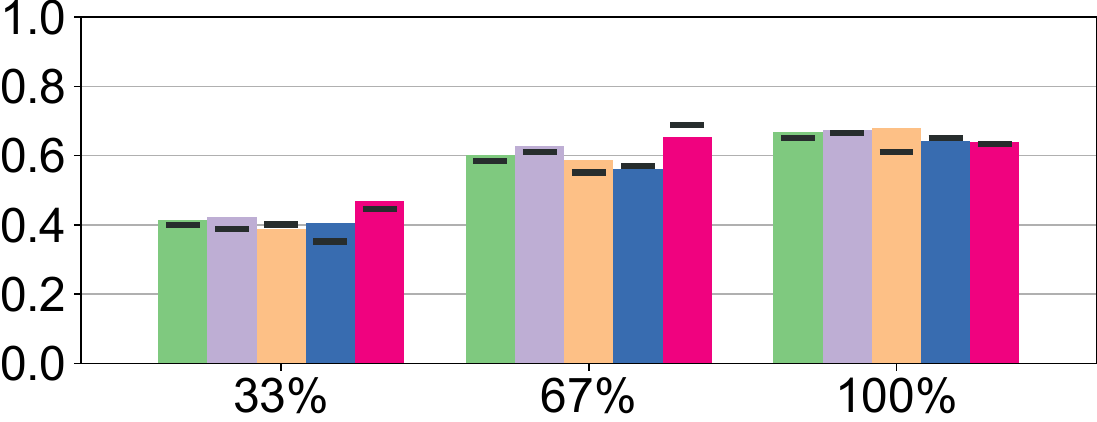}
         \caption{ACIDS Shallow}
         \label{fig:snn_general_performance_ACIDS_accuracy}
     \end{subfigure}
     \hfill
     \begin{subfigure}[t]{0.31\textwidth}
         \centering
         \includegraphics[width=\textwidth]{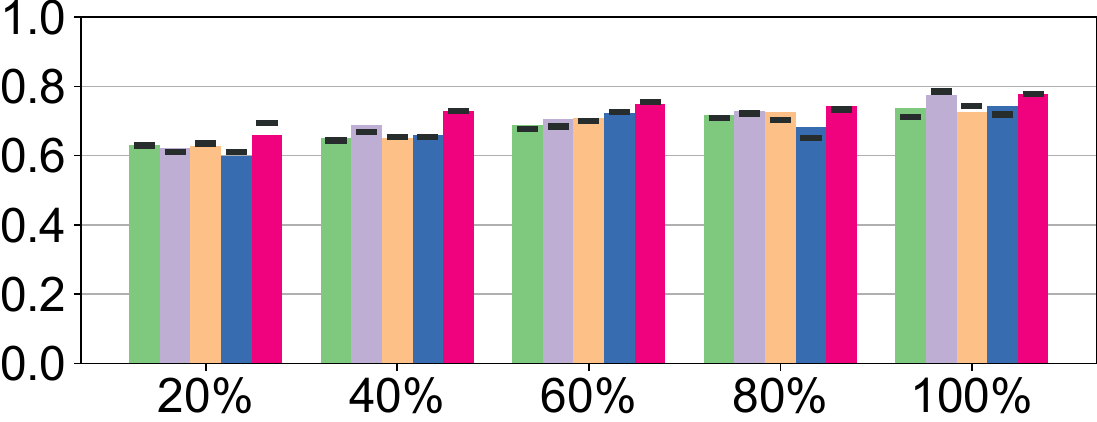}     
        \caption{Wearable-HAR Shallow}
        \label{fig:snn_general_performance_Wearable-HAR_accuracy}
     \end{subfigure}
     \hfill
     \begin{subfigure}[t]{0.31\textwidth}
         \centering
         \includegraphics[width=\textwidth]{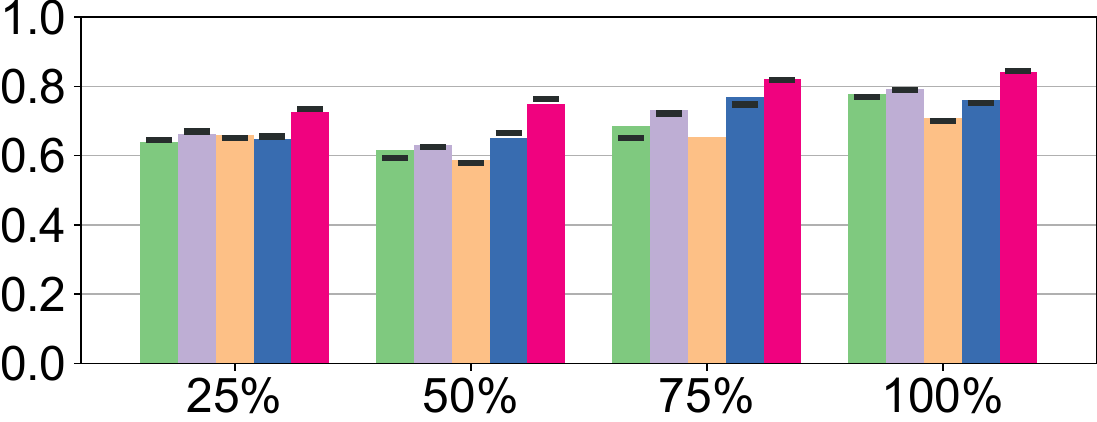}
        \caption{Wi-Fi-HAR Shallow}
         \label{fig:snn_general_performance_Wi-Fi-HAR_accuracy}
     \end{subfigure}
     \hfill
     
    % DeepSense as downstream
     \begin{subfigure}[b]{0.31\textwidth}
         \centering 
         \includegraphics[width=\textwidth]{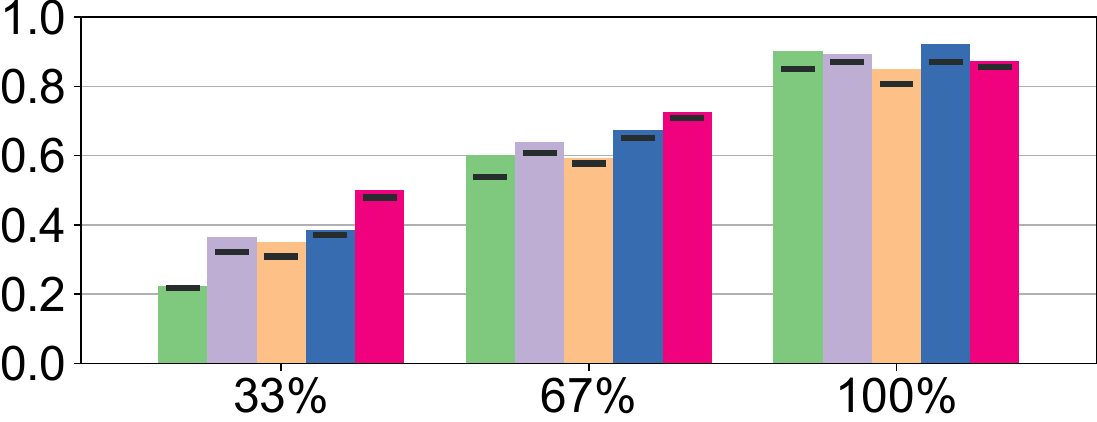}
         \caption{ACIDS DeepSense}
         \label{fig:deepsense_general_performance_ACIDS_accuracy}
     \end{subfigure}
     \hfill
     \begin{subfigure}[b]{0.31\textwidth}
         \centering
         \includegraphics[width=\textwidth]{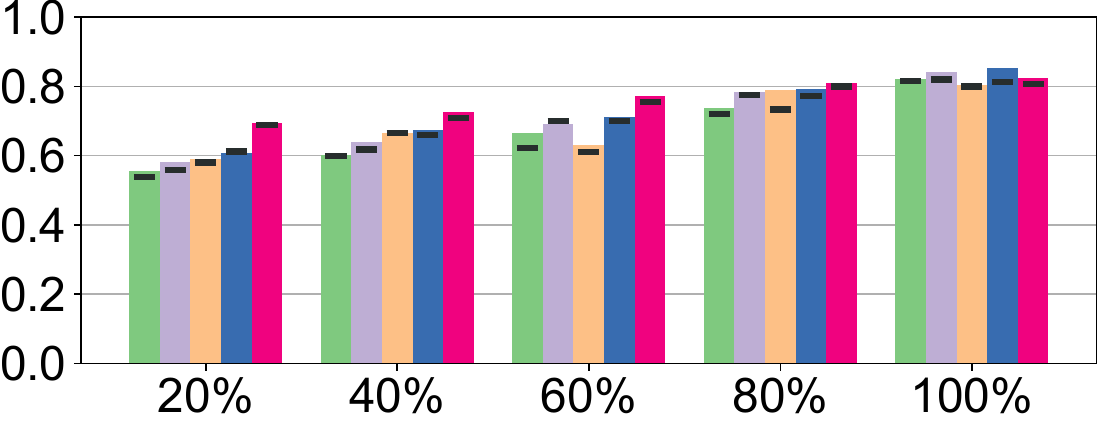}     
        \caption{Wearable-HAR DeepSense}
        \label{fig:deepsense_general_performance_Wearable-HAR_accuracy}
     \end{subfigure}
     \hfill
     \begin{subfigure}[b]{0.31\textwidth}
         \centering
         \includegraphics[width=\textwidth]{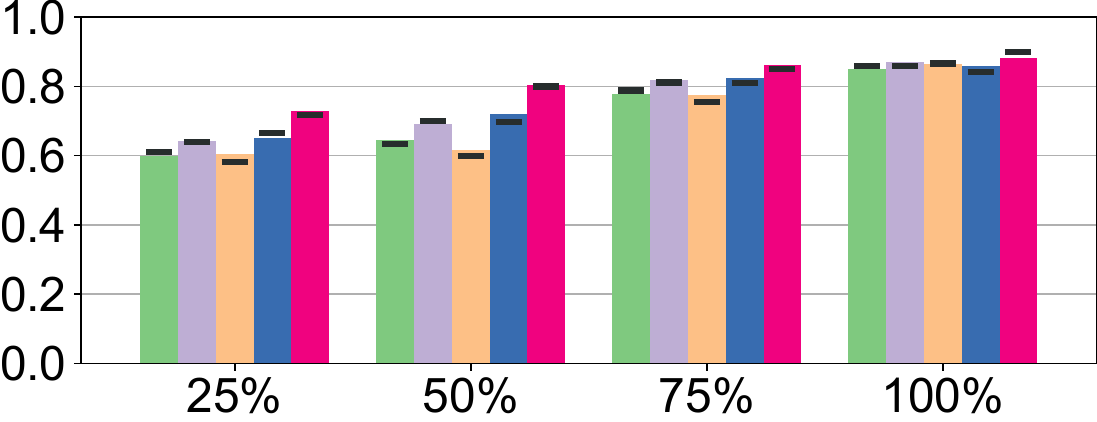}
        \caption{Wi-Fi-HAR DeepSense}
         \label{fig:deepsense_general_performance_Wi-Fi-HAR_accuracy}
     \end{subfigure}
     \hfill

    % Transformer as downstream
     \begin{subfigure}[b]{0.31\textwidth}
         \centering 
         \includegraphics[width=\textwidth]{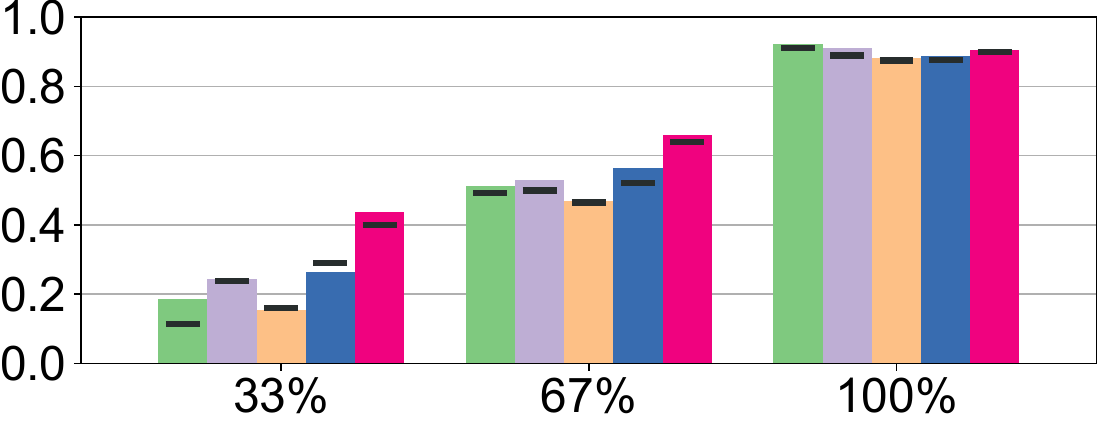}
         \caption{ACIDS Transformer}
         \label{fig:transformer_general_performance_ACIDS_accuracy}
     \end{subfigure}
     \hfill
     \begin{subfigure}[b]{0.31\textwidth}
         \centering
         \includegraphics[width=\textwidth]{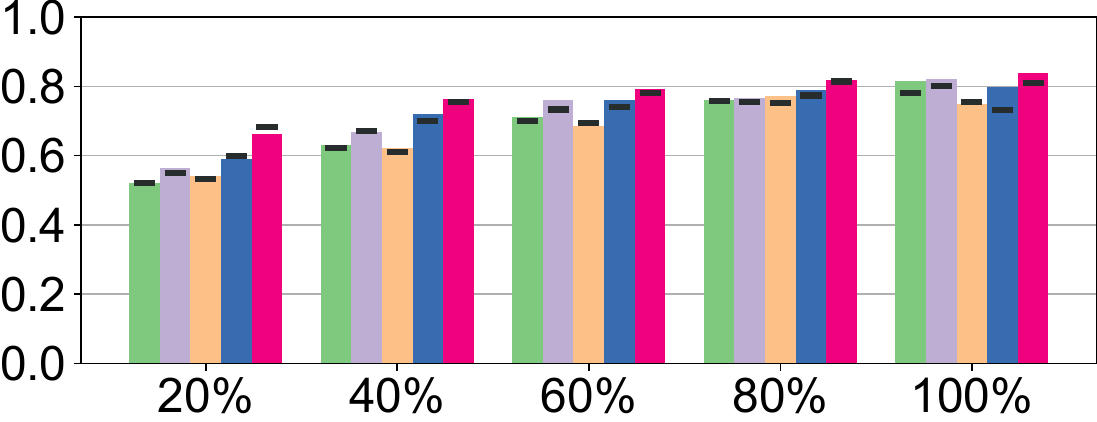}     
        \caption{Wearable-HAR Transformer}
        \label{fig:transformer_general_performance_Wearable-HAR_accuracy}
     \end{subfigure}
     \hfill
     \begin{subfigure}[b]{0.31\textwidth}
         \centering
         \includegraphics[width=\textwidth]{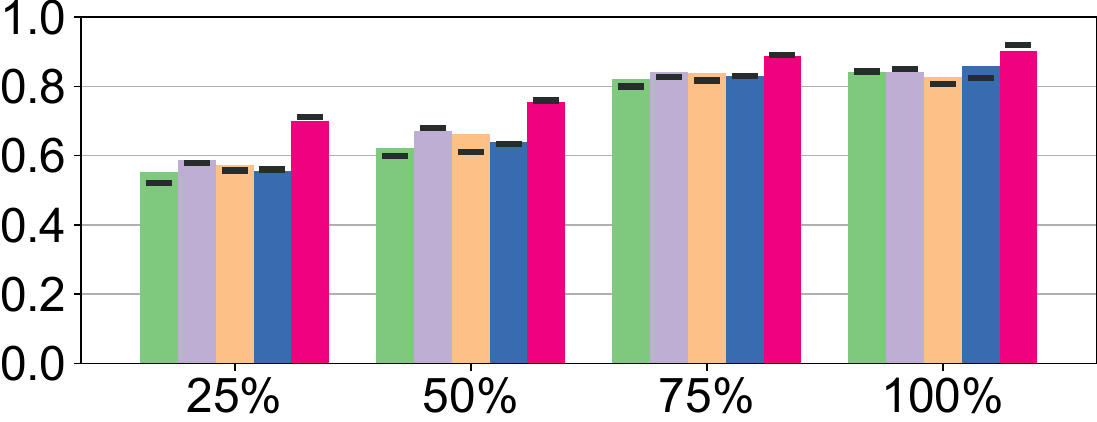}
        \caption{Wi-Fi-HAR Transformer}
         \label{fig:transformer_general_performance_Wi-Fi-HAR_accuracy}
     \end{subfigure}
     \hfill
      
    \caption{Each row of figures presents results from a particular type of downstream classifier, while each column of figures shows results from a specific dataset. Height of each bar represents the accuracy value while the vertical position of each dash indicates the F1-score. The percentages on the x-axis represent the proportion of filled cells within the Sudoku matrix. A higher percentage coverage indicates a more complete dataset.}
        \label{fig:dataset_performance}
\end{figure*}

\subsection{Baselines}
\label{sec:base}
We select four comparison baselines (alternative approaches to improve training robustness) to understand the advantages of SudokuSens in handling IoT data scarcity challenges, compared to feasible alternatives.

\begin{itemize}
\item
\textbf{Basic}.
% \subsubsection{Basic}
We train the downstream neural network on the original dataset by supervised learning. This is the most straightforward and common approach taken in many IoT applications. It offers no robustness support.
\item 
\textbf{Conventional data augmentation}.
% \subsubsection{Conventional data augmentation} 
Data augmentation is a common approach to enhance the diversity and size of datasets and increase the robustness of trained models.
We adopt the data augmentation techniques used in \cite{zhang2022self}, which include both augmentation in time and frequency domain.
We randomly apply one time-domain augmentation and one frequency-domain augmentation on each sample and create 10 augmented samples for each original.
Then we use supervised learning to train the downstream neural network.
\item
\textbf{Temporal Neighborhood Coding (TNC)}.
% \subsubsection{Temporal Neighborhood Coding (TNC)}
TNC is a temporal contrastive learning framework designed to capture the progression of the underlying temporal dynamics~\cite{tonekaboni2021unsupervised}.
The idea is to divide the temporal sequence into windows and define a neighborhood around every window.
Then representations are learned by contrasting samples from the same neighborhood and samples from different neighborhoods.
Comparing with our approach, TNC provides a temporal contrastive learning baseline that is not session-aware.
\item
\textbf{Time-Frequency Consistency (TF-C)}.
% \subsubsection{Time-Frequency Consistency (TF-C)} 
TF-C is a state-of-the-art general contrastive learning framework for time-series data~\cite{zhang2022self}.
It promotes proximity of time and frequency-based representations of identical time series samples in the latent space while distancing those from different samples.
This process incorporates both data augmentation and contrastive learning.
Thus, TF-C serves as a strong baseline in our study.
\end{itemize}

\subsection{Overall Performance on Datasets}
\label{subsec:overall_performance}
Figure~\ref{fig:dataset_performance} presents SudokuSens and baseline performance on each dataset with different classifiers. 
The {\em percentage coverage\/} of the Sudoku matrix is varied as shown on the $x$-axis. Smaller coverage values indicate higher data scarcity. When SudokuSens is used, each of the empty cells, $\mathcal{C}_{unseen}$, of the matrix is filled-in with synthetic traces whose length matches the {\em average\/} number of samples per cell for the covered cells, $\mathcal{C}_{seen}$.

Several observations from the figure are worth commenting on. First, in general, SudokuSens improves classification accuracy over other robustness/augmentation baselines. The difference becomes more pronounced as the percentage coverage by training/validation data decreases. At 100\% coverage, SudokuSens does not offer a significant advantage, if any.

Second, more advanced classifiers (lower rows in Figure~\ref{fig:dataset_performance}), appear to be less robust. While they offer better accuracy and F1-score at 100\% coverage (no domain shift between training and testing), they fare worse than simpler classifiers at low coverage values. In other words, their degradation is more abrupt as data scarcity increases. Figure~\ref{fig:dataset_performance} shows that SudokuSens generally offers a better advantage at lower coverage values and with more advanced (and thus less robust) classifiers.  

\begin{figure}[h]
  \centering
  \subfloat{
    \includegraphics[width=0.20\textwidth]{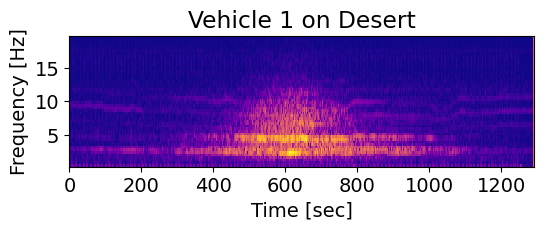}
  }
  % \hfill  % This will add horizontal space between the subfigures
  \subfloat{
    \includegraphics[width=0.20\textwidth]{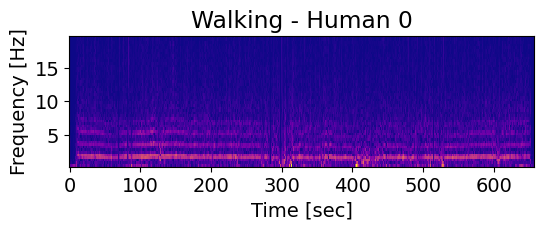}
  }
  \\
  \subfloat{
    \includegraphics[width=0.20\textwidth]{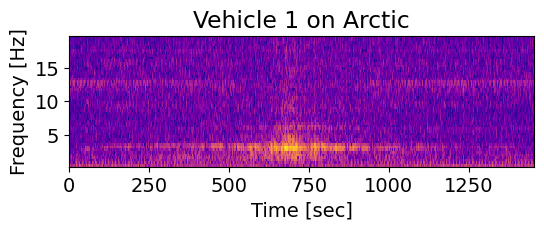}
  }
  % \hfill  % This will add horizontal space between the subfigures
  \subfloat{
    \includegraphics[width=0.20\textwidth]{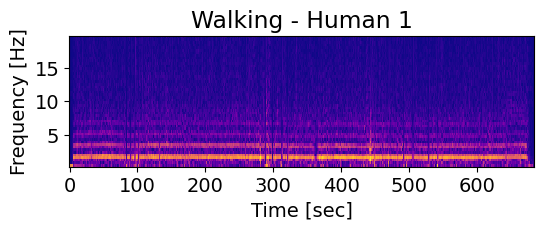}
  }
  \\
  \subfloat{
    \includegraphics[width=0.20\textwidth]{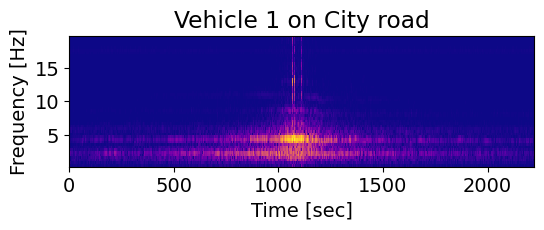}
  }
  % \hfill  % This will add horizontal space between the subfigures
  \subfloat{
    \includegraphics[width=0.20\textwidth]{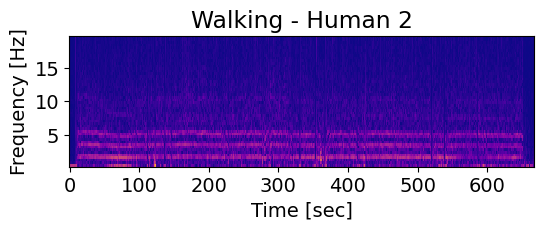}
  }
\caption{The figure shows that similarity in target signatures across different terrains (in ACIDS data) is much less than similarity in
activity signature across different subjects (in Wearable HAR data). SudokuSens offers a higher value from data interpolation in the ACIDS dataset.}
\label{fig:similarity}
\end{figure}

Another interesting observation is the difference in SudokuSens benefits across data sets (i.e., across the columns in Figure~\ref{fig:dataset_performance}). The figure demonstrates a higher improvement due to SudokuSens on the ACIDS vehicle detection dataset (leftmost column) compared to the two HAR datasets. The comparison offers insights into data set properties that might be more conducive to improvements with SudokuSens. This observation needs some elaboration:
SudokuSens is designed to improve inference primarily by generating new training traces for additional environmental conditions (via data extrapolation using the CVAE). Clearly, a main factor affecting the utility of such data extrapolation is the degree of similarity between seen and unseen conditions. If the unseen conditions for a target class are similar to the ones already represented in the training data (for that class), then the improvement attained from the extrapolation is marginal. Otherwise, the extrapolation is of more value. This explains the difference between the columns. The sensory signatures of the same vehicle in different terrains are quite distinct. Thus, SudokuSens extrapolation to mitigate the large domain shift is advantageous. In contrast, the signature variability in performing the same activity (such as walking) across individuals is less pronounced. Thus, extrapolation has less value. Figure~\ref{fig:similarity} visually confirms the above by comparing example sensory signatures for the same class of output in Wearable-HAR versus ACIDS. It can be seen from Figure~\ref{fig:similarity} that the differences between spectrograms in the left column (seismic signatures of the same vehicle in different terrains) are bigger than the differences between spectrograms in the right column (accelerometer signatures of the same activity for different people). Thus, SudokuSens offers more value from its extrapolation framework in the ACIDS dataset.

Finally, observe from Figure~\ref{fig:dataset_performance} that at/near 100\% coverage, SudokuSens offers a marginally better advantage over baselines for the HAR datasets compared to ACIDS (where it offers no advantage at all). At such a high coverage, CVAE extrapolation is not a factor. Instead, performance differences are attributed to the used disturbance rejection framework (i.e., the SA-TCL encoder). Note that, at fine-grained timescales, sensory signatures of human activities are temporally more complex than the signatures of a rotating vehicle engine. Thus, there is more value in disentangling activity signatures from dynamic background disturbances in the HAR dataset, compared to the ACIDS dataset. In ACIDS, simpler techniques are sufficient for disturbance rejection, which is why, for ACIDS, SudokuSens does not beat the baselines (at 100\% coverage).

\subsection{Outdoor Field Experiments}
\label{subsec:case_study_performance}
\begin{figure}[h]
  \centering
  \subfloat[The sensor node: RPi 4 integrated with a seismic sensor and a microphone array]{
    \includegraphics[width=0.19\textwidth]{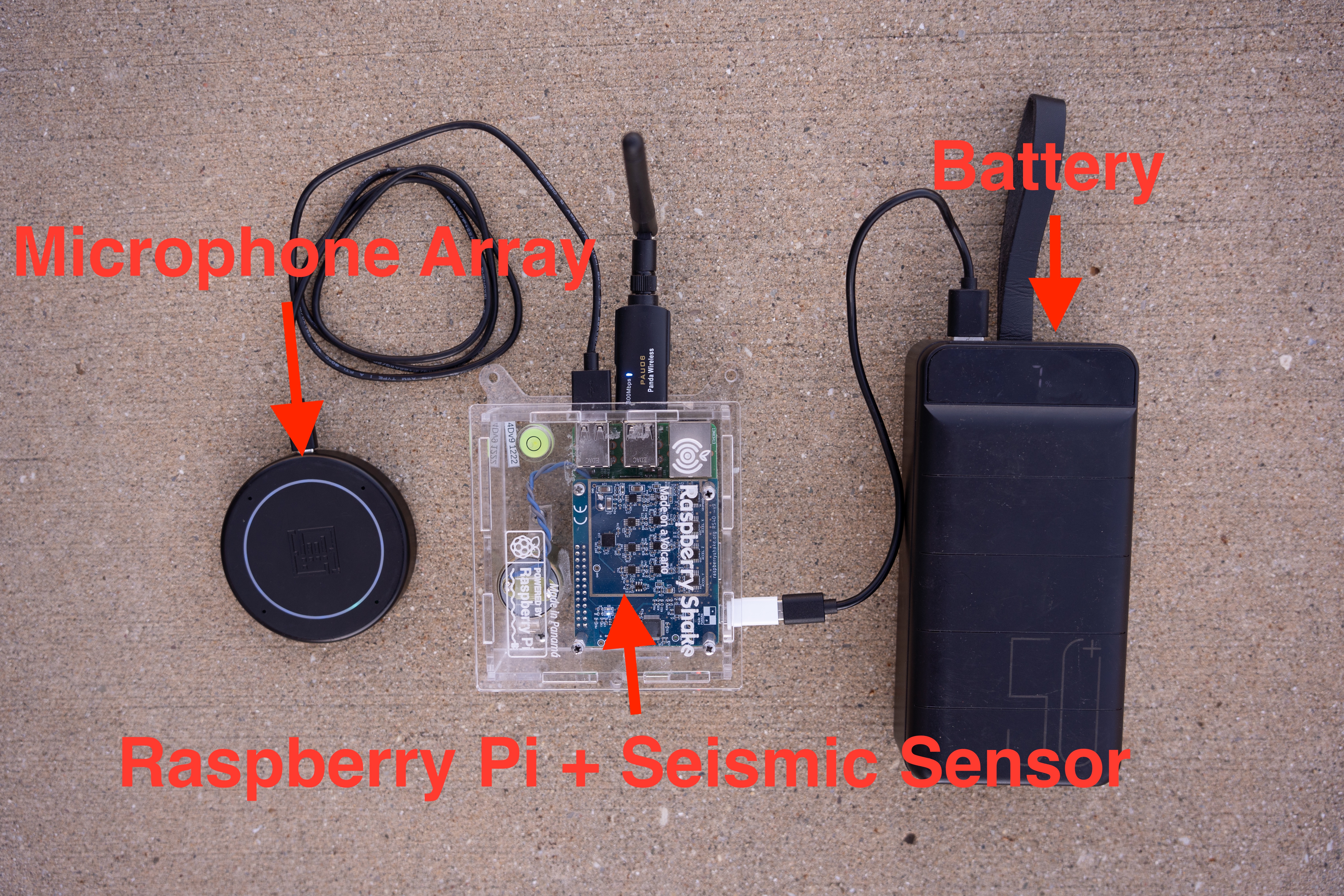}
    \label{fig:raspberryshake}
  }
  \hfill  % This will add horizontal space between the subfigures
  \subfloat[A target: a Ford Mustang sports car]{
    \includegraphics[width=0.19\textwidth]{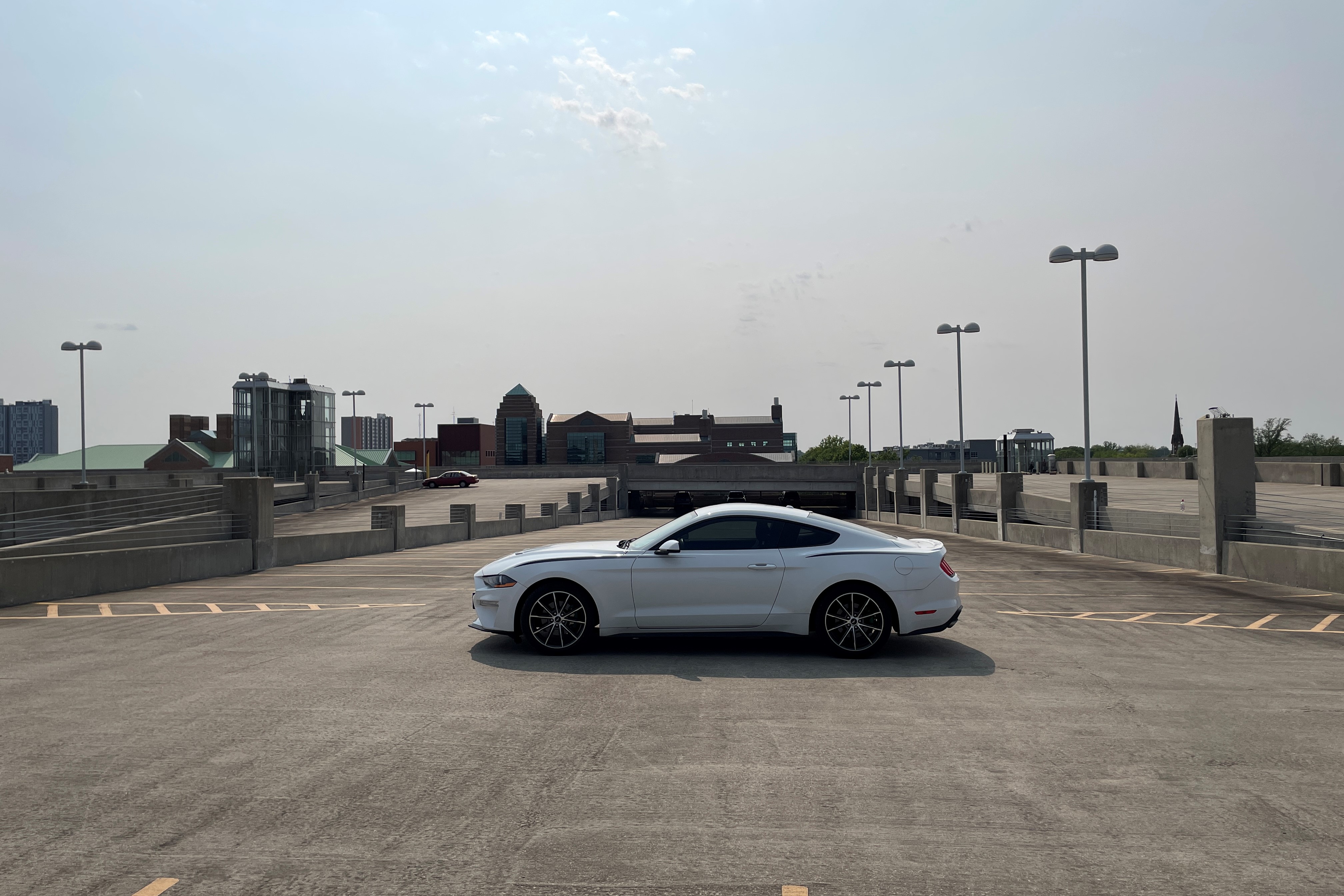}
    \label{fig:edge}
  }
  \caption{Hardware devices adopted in the case studies.}
  \label{fig:hardware}
\end{figure}

To further confirm the insights discussed above, we conduct additional outdoor experiments featuring a deployed run-time system executing on the target edge hardware. This system has already been mentioned briefly in Section~\ref{sec:motivation}. We describe it below in more detail.
Figure~\ref{fig:hardware} illustrates the sensor devices used. Namely, sensing is done with a RaspberryShake 1D~\cite{RaspberryShake}, which is a Raspberry Pi~4 connected to a seismic sensor. Raspberry Pi 4 is a compact and cost-effective mobile computer, equipped with a 1.5GHz Cortex A72 CPU and 2 GB of memory. In addition, a ReSpeaker USB Mic Array~\cite{mic_array} was connected to the Raspberry Pi 4 via a USB port. It has 4 high-performance digital microphones and supports 360-degree audio pick-up. The sensor devices collect seismic and acoustic signals at a sampling rate of 100 Hz and 8000 Hz respectively. The entire node is situated on firm ground as depicted in Figure~\ref{fig:raspberryshake}. A USB Wi-Fi antenna was connected to the Raspberry Pi 4, allowing it to be remotely controlled for experimental purposes by our remote controller device: a Lenovo ThinkPad T430 laptop (shown in Figure~\ref{fig:edge}). The laptop was {\em not\/} used to process the inference workload. All inference and detection were deployed on the Raspberry Pi, and performed in real-time on successive 2-second sensor data intervals.

%During {\em training\/}, sensors were activated remotely by the laptop, and data were collected, uploaded, then used for training. 
%All inference-time code was deployed on the Raspberry Pi 4, where it .
%The detected target class, along with the timestamp, are logged locally.
While operating the target vehicle, the driver carried a smartphone to record GPS traces. For evaluation (i.e., ground-truthing) purposes, 
when the distance between the vehicle and the sensor exceeded 100 ft, we considered that there was ``no vehicle" nearby.
Conversely, when the vehicle was within 100 ft, the ground truth vehicle type was recorded.

SudokuSens was pretrained with the conditions listed in Table~\ref{tab:sampled_conditions} (where scenarios A through G are as defined in  Table~\ref{tab:scenarios}).
We then filled out the rest of the SudokuSens matrix (i.e., the empty slots in Table~\ref{tab:sampled_conditions}).
Collection scenarios A through G were treated as {\em confounding attributes\/} for each target class. They varied in such properties as the nature of the experimental location (e.g., urban versus rural) and the type of underlying terrain (e.g., paved versus gravel). 
Each scenario (A through G) featured additional internal variability arising from dynamically changing target speed, distance from sensor, and the naturally occurring background noise, forming acoustic and seismic {\em dynamic disturbances\/}.
While SudokuSens is executed in real time, the performance of other baselines was evaluated via data playback.
%The first stage of SudokuSens is conditional interpolation on the intrinsic attributes. During training, we feed the data samples and their corresponding condition vectors, which include the embeddings of vehicle type and location, into the CVAE. After training, the decoder of the CVAE takes in the missing ``vehicle-location'' combinations, thereby directing the neural network to interpolate missing conditions tied to the empty cells in Table~\ref{tab:sampled_conditions}. 
%This process includes the generation of samples for Ford Mustang in scenario D, which are particularly beneficial in improving classifier performance in case study 1. 
%The generated samples, together with the original dataset, are input into the session-aware temporal contrastive learning. 
%During this process, the encoder learns how to mitigate variances from dynamic disturbances, including the unseen wind noise variations in case study 2. 
%The trained encoder is then integrated with the downstream classifier. Finally, we deploy SudokuSens together with the downstream classifier onto the Raspberry Pi 4 and evaluate its performance. 

\begin{figure}[h]
  \centering
  \subfloat[Unseen condition 1]{
    \includegraphics[width=0.22\textwidth]{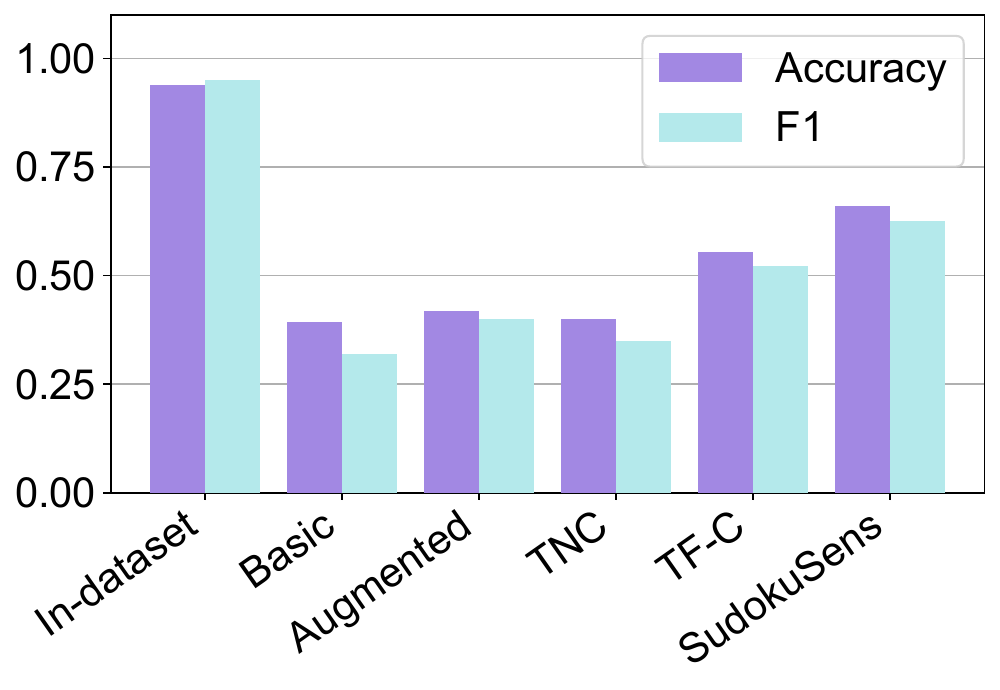}
    \label{fig:case_study_eval_case1}
  }
  \hfill  % This will add horizontal space between the subfigures
  \subfloat[Unseen sondition 2]{
    \includegraphics[width=0.22\textwidth]{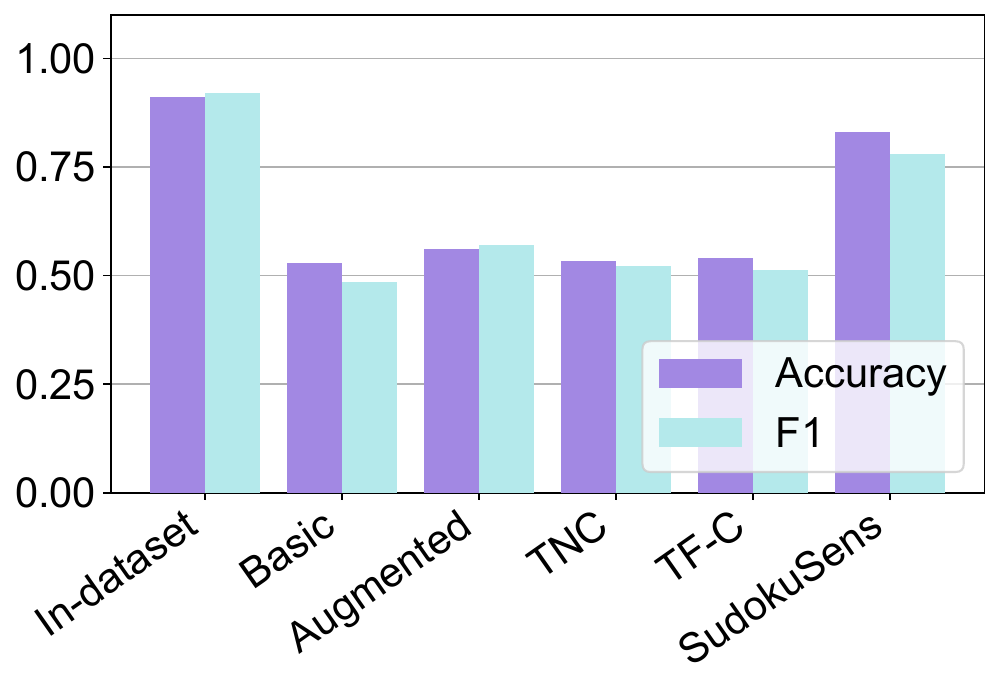}
    \label{fig:case_study_eval_case2}
  }
  \caption{SudokuSens experimental study}
  \label{fig:sudokusens_case_study}
\end{figure}

Figure~\ref{fig:sudokusens_case_study} compares the performance of SudokuSens to the baselines listed in Section~\ref{sec:base} in terms of classification accuracy and F1 score in the case of {\em unseen condition~1\/} and {\em unseen condition~2\/} (described in Section~\ref{subsec:case_study1} and Section~\ref{subsec:case_study2}, respectively), as well as the case where no unseen conditions exist (denoted \textit{in-dataset}). The latter serves as the upper limit of model performance. The figure presents additional empirical evidence showing that SudokuSens outperforms the other baselines.

\subsection{Generalizability and Limits}
\label{subsec:generalizability}

The above results show great improvements brought by SudokuSens for various downstream classification tasks. 
When do these improvements stop and what do they depend on? 
Below, we investigate this question by conducting three additional experiments, aiming to understand what factors impact the efficacy of SudokuSens. 
% In the first 2 experiments, we manipulate the sampling sparsity and temporal variance that exist in the vehicle detection dataset described in Section~\ref{sec:motivation}. 
% By comparing the outcome with the results from Section~\ref{sec:motivation}, we give a more principled understanding of the true capabilities and limitations of the proposed approaches.
% In the 3rd experiment, we incorporate GPS trace data into the vehicle detection dataset described in Section~\ref{sec:motivation}, converting the task to predict the distance between the vehicle and the sensor, which is a continuous output variable. 
% We show that SudokuSens can successfully generalize to this regression task.

\subsubsection{Limits of CVAE Extrapolation Efficacy}
\begin{figure}[h]
  \centering
  \subfloat[Marginal knowledge transfer]{
    \includegraphics[width=0.22\textwidth]{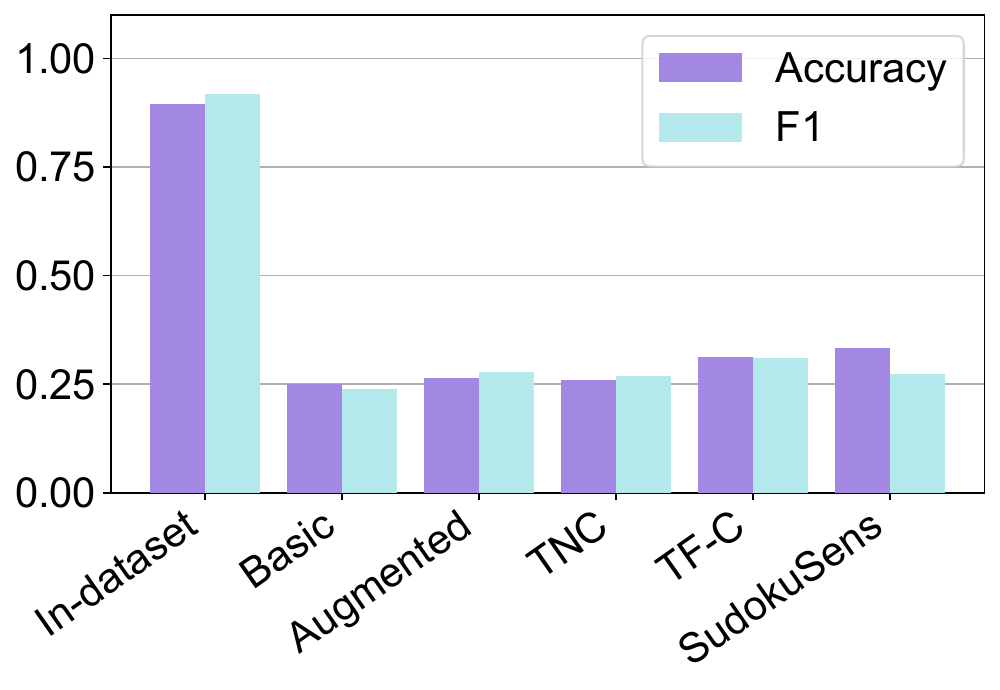}
    \label{fig:conditional_interpolation_variantA}
  }
  \hfill  % This will add horizontal space between the subfigures
  \subfloat[No knowledge transfer]{
    \includegraphics[width=0.22\textwidth]{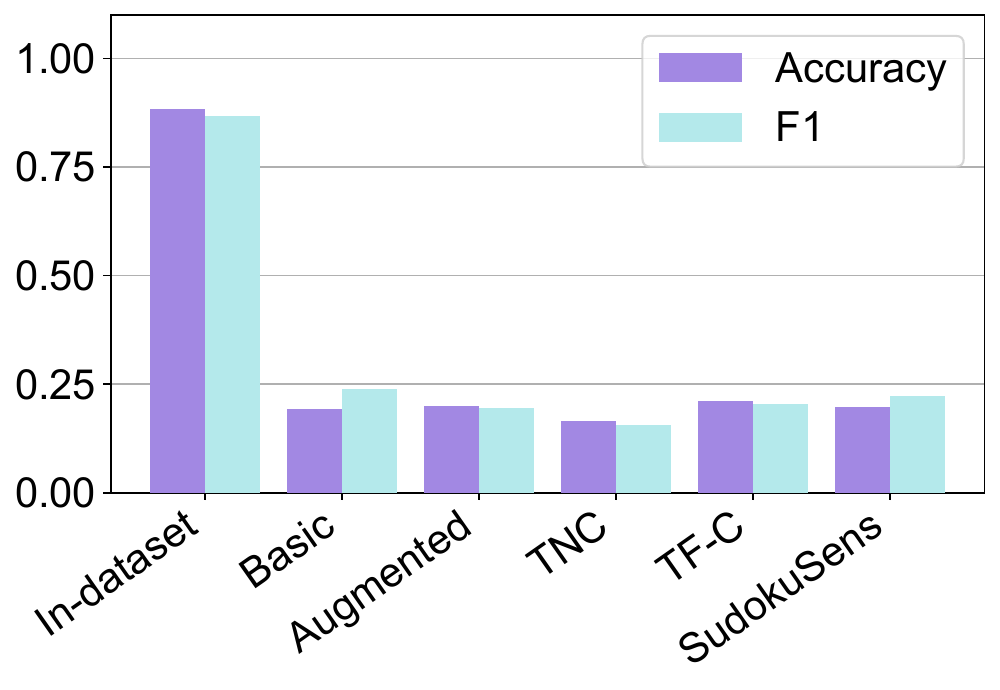}
    \label{fig:conditional_interpolation_variantB}
  }
  \caption{Generalizability of conditional interpolation.}
  \label{fig:generalizability_conditional_interpolation}
\end{figure}

The main factor affecting the quality and efficacy of data extrapolation using the CVAE is the sparsity of coverage in the training data. For a given target, classification accuracy in new conditions depends on how many $\mathcal{C}_{seen}$ cells of the Sudoku matrix features that target in the first place. To confirm this intuition, 
we consider the settings of {\em unseen condition~1\/} described in Section~\ref{subsec:case_study1} (i.e., the Ford Mustang running in scenario F), and gradually reduce the number of seen “cells”, $\mathcal{C}_{seen}$, in the Sudoku matrix shown in Figure~\ref{tab:sampled_conditions} that feature the Ford Mustang. We then apply SudokuSens and other baselines to the reduced data set. Figure~\ref{fig:generalizability_conditional_interpolation} shows the Ford Mustang classification accuracy and F1 score in two such cases. One when the Mustang data in scenarios A and C was removed from the training data (left), and one where the Mustang data in scenarios A, B, and C was removed. Note how, in the left figure, SudokuSens advantage is marginal, whereas in the right it offers no advantage. This is because removing scenarios A, B, and C leaves no row or column overlap between Ford Mustang observations and other $\mathcal{C}_{seen}$ data. Thus, knowledge is not transferred from seen Mustang cells to other cells. Similarly, in the right figure, the transfer is minimal. We conjecture that CVAE extrapolation works when at least one rectangle can be found in the Sudoku matrix such that three of its corners are cells in $\mathcal{C}_{seen}$ and the fourth is the unknown condition we want to classify the target in. The more such rectangles exist that involve the unknown condition, the better the knowledge transfer.

\subsubsection{Limits of SA-TCL Disturbance Rejection}
\begin{figure}[h]
  \centering
  \subfloat[Top 50\%]{
    \includegraphics[width=0.22\textwidth]{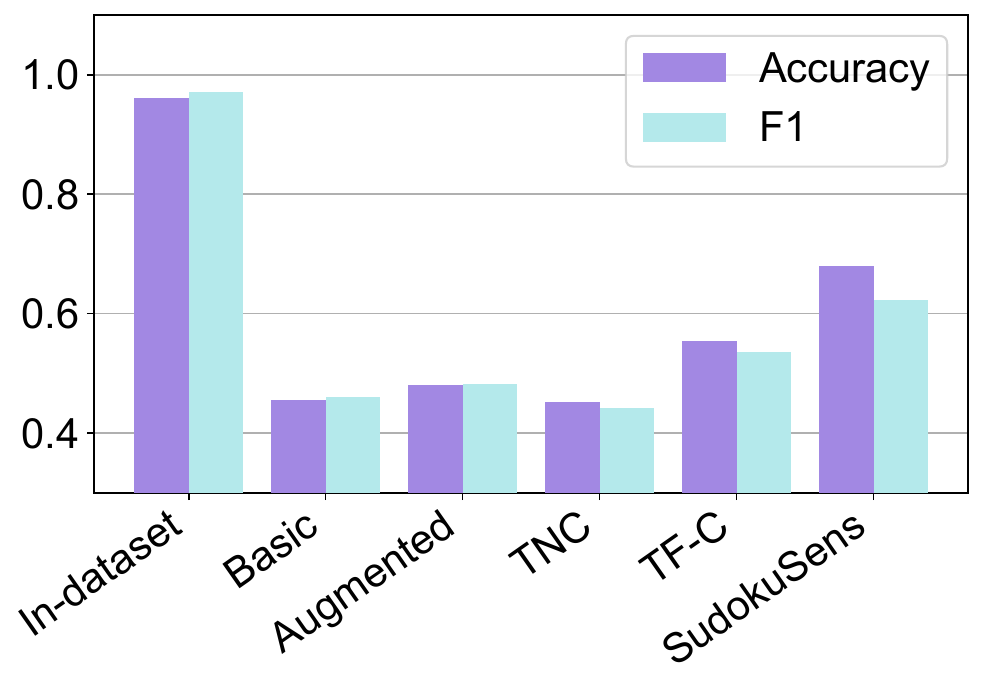}
    \label{fig:satcl_0.5}
  }
  \hfill  % This will add horizontal space between the subfigures
  \subfloat[Top 25\%]{
    \includegraphics[width=0.22\textwidth]{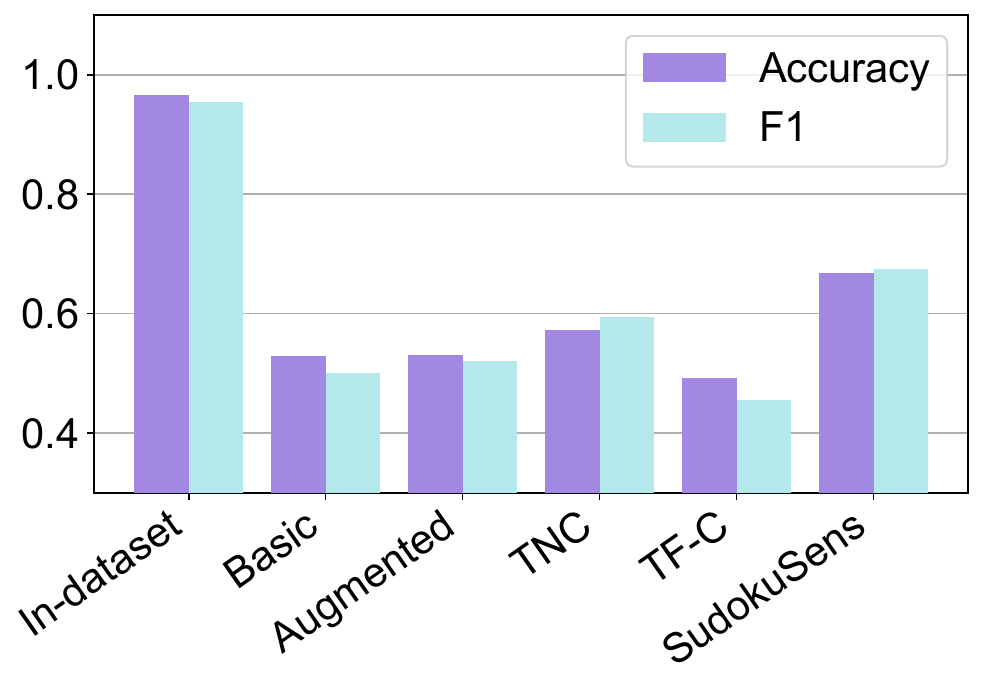}
    \label{fig:satcl_0.25}
  }
  \vspace{-0.1in}
  \caption{Generalizability of SA-TCL. As the temporal dynamics in the dataset decrease, SudokuSens exhibits less superiority compared to the baselines.}
  \label{fig:generalizability_satcl}
  \vspace{-0.1in}
\end{figure}

Aiming at reducing label sensitivity to dynamic disturbances, SA-TCL helps reduce susceptibility to noise. 
A higher level of disturbances and other data variability within individual session traces provides greater opportunities for SA-TCL to enhance the performance of downstream tasks. 
Conversely, if the dataset exhibits limited temporal variability within individual session traces, the improvement offered by SA-TCL is constrained.   
% Meanwhile, it may map diverse parts of the same session closer together in the latent space making it harder for a downstream classifier to recognize the actual temporal activity. 
To confirm this intuition, we perform additional experiments under the settings of {\em unseen conditions~1\/} described in Section~\ref{subsec:case_study1}, except for filtering out data samples within the same session by their amplitude to cut down temporal variability. 
We show two experiments where we only keep the samples that fall within 50\% and 25\% difference from the max seismic sample energy, respectively. Keeping only higher energy samples reduces data variability, thus limiting the influence of SA-TCL.
Figure~\ref{fig:generalizability_satcl} shows the results. It indicates that a reduction in temporal variability within sessions (see right figure) leads to a decrease in the benefits derived from SudokuSens. Other baselines do better, thereby eroding the advantage of SudokuSens. In contrast, the left figure shows a better advantage over baselines.

\subsubsection{Generalizability to Regression Tasks}
\begin{figure}[h]
\centering
\includegraphics[width=0.8\linewidth]{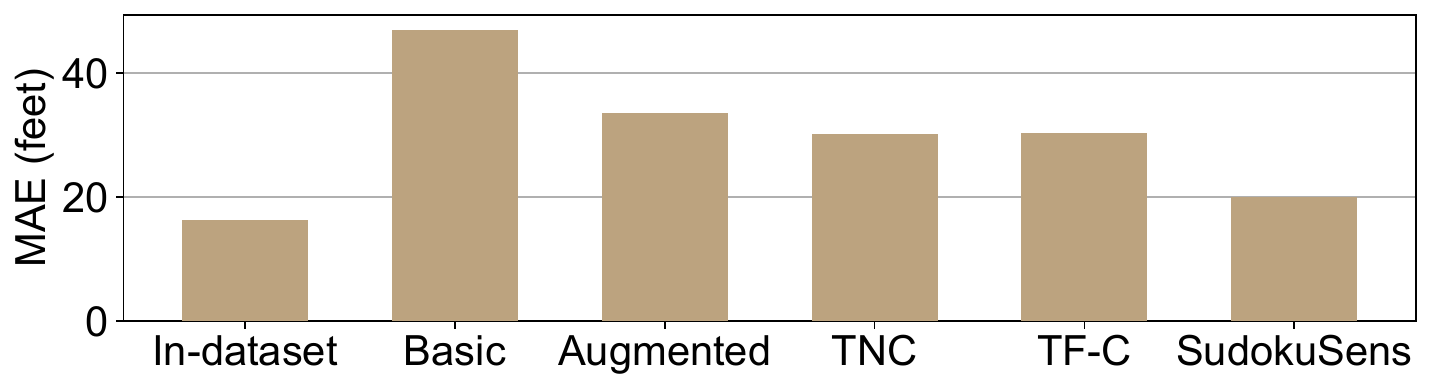}
\caption{SudokuSens in distance prediction task}
\label{fig:distance_regression}
\end{figure}

While the paper thus far focused on classification tasks, in this section, we show that SudokuSens improves the performance of regression tasks as well.  
In this experiment, we replace the classifier with a neural network trained to perform a regression task based on the same vehicle detection dataset used in Section~\ref{subsec:case_study1}, but with GPS traces of the vehicles included to train the regression algorithm. 
In this task, the target is to predict distance from the vehicle to the sensor. 
Like in vehicle detection, we segment the whole session of sensing data and the GPS trace into 2-second chunks. 
The sampling rate of GPS is 1Hz and we take the 2nd second of GPS coordinate in each chunk to calculate the ground truth distance to the sensor location. 
As for the downstream neural network architecture, we use a version of DeepSense designed for regression tasks.
We use the mean squared error (MSE) as the loss function during training.
We evaluate the improvement in distance estimation accuracy due to SudokuSens. The results are shown in Figure~\ref{fig:distance_regression}.
The figure shows that SudokuSens enables a more accurate distance prediction than the baselines for the unseen environmental condition under investigation. While the error (20 ft) might seem large, it is actually of the same order of magnitude as the underlying GPS ground truth. Also, since the samples are two seconds long, additional labeling inaccuracy occurs due to target motion within a window.
 
\subsection{Ablation Studies}
Next, we conduct an ablation study on the individual components in SudokuSens, investigating their respective contributions to the overall model performance.

\subsubsection{Conditional Interpolation}
The CVAE in SudokuSens
uses a generative approach to create additional data in $\mathcal{C}_{unseen}$ cells. The default considered thus far was to populate those cells with the same amount of data on average as the $\mathcal{C}_{seen}$ cells. What if that ratio was changed? How much data would be too much? How much is not enough? To explore this question, 
we set up the ablation study for conditional interpolation by changing the interpolation ratio, which we define as the ratio of interpolated samples (synthesized from CVAE) to the average number of samples per cell in the original dataset. An interpolation ratio of 0 means that we remove conditional interpolation from the framework, and only apply SA-TCL to the original dataset.

\input{content/tables/ablation_condinterp}

The results for the data sets from Section~\ref{sec:dataset} are shown in Table~\ref{tab:ablation_study_condinterp}. While conditional interpolation contributes to performance improvements across all interpolation ratios in this table (compared to results without interpolation), there is indeed a sweet spot at which the benefit is maximized. Generally, shallow neural networks tend to yield optimal performance at a lower interpolation ratio (around 1x), while larger classifiers seem to perform better with higher interpolation ratios. This is plausible since their model complexity can benefit from the additional data. We conjecture that, eventually, a high enough ratio may cause performance degradation, but the table focuses on the sweet spot at which benefits are maximized. Similar observations were seen for the outdoor study. They are not shown due to space limitations.

%\textcolor{blue}{As for the profiling results for the case studies, the ideal interpolation ratio decreases when the number of intersected conditions drops (as in case study 1 variant $\alpha$). The synthetic samples even impose a negative effects on the performance when there is no intersected condition in the original dataset (as in case study 1 variant $\beta$). This demonstrates that conditional interpolation relies on the intersections of attributes in the "Sudoku table" to realize knowledge transfer. If the path of knowledge transfer is cut off, the generated samples are not realistic enough to enhance the downstream task, and even potentially damage the performance.}

%\textcolor{blue}{When the data volume decreases, the requirement for synthetic data also diminishes, as shown in the results from the filtered datasets of case study 1.  The realness of the synthetic data from conditional interpolation relies on the realistic data from the original dataset. Given fewer realistic data, the capability of conditional interpolation in generating representative samples is weaker. Thus, the ideal interpolation ratio get smaller.}

\subsubsection{Session-Aware Temporal Contrastive Learning}
Next, we show a performance comparison with and without SA-TCL when the interpolation ratio is 1x in Table~\ref{tab:ablation_study_satcl}. SA-TCL is shown to consistently contribute a performance improvement across all datasets and different downstream classifiers.

\input{content/tables/ablation_sa-tcl}

\begin{figure}
  \centering
  \subfloat[Before SA-TCL, \\ SC=-0.28]{
    \includegraphics[trim=0 35 10 45,clip,width=0.20\textwidth]{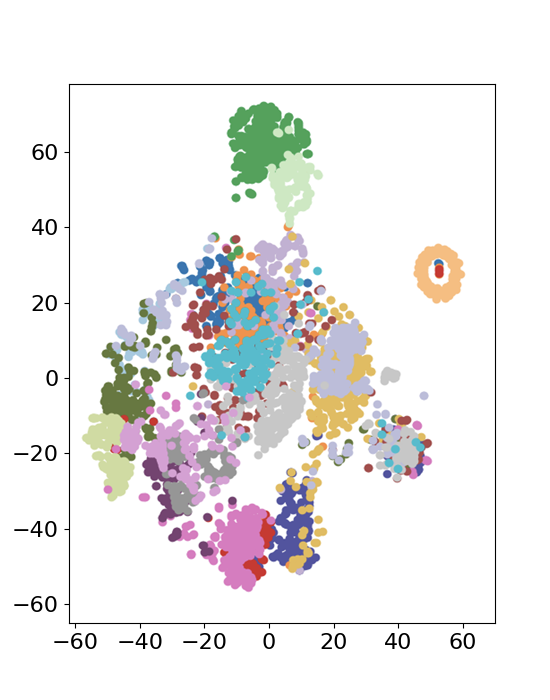}
    \label{fig:before_sa-tcl_acids}
  }
  % \hfill  % This will add horizontal space between the subfigures
\hspace{0.5cm}
  \subfloat[After SA-TCL, \\ SC=0.40]{
    \includegraphics[trim=0 35 10 45,clip,width=0.20\textwidth]{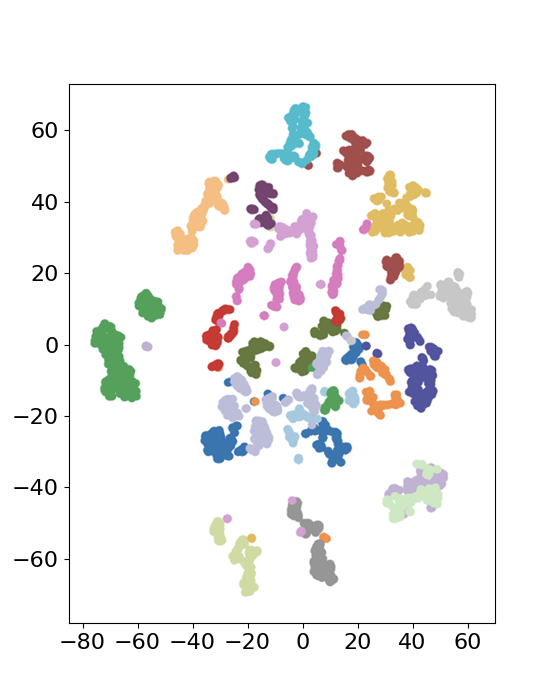}
    \label{fig:after_sa-tcl_acids}
  }

  % \subfloat[Wearable-HAR before SA-TCL, SC=0.22]{
  %   \includegraphics[width=0.15\textwidth]{figure/evaluation/tempclr_visualization/before_tempclr_har.png}
  %   \label{fig:before_sa-tcl_har}
  % }
  % % \hfill  % This will add horizontal space between the subfigures
  % \subfloat[Wearable-HAR after SA-TCL, SC=0.49]{
  %   \includegraphics[width=0.147\textwidth]{figure/evaluation/tempclr_visualization/after_tempclr_har.png}
  %   \label{fig:after_sa-tcl_har}
  % }
  
  \caption{Visualization of the inputs and the output embeddings from SA-TCL. Each color represents one data collection session in ACIDS.}
  \label{fig:before_and_after_sa-tcl}
\end{figure}

To intuitively understand its effect, we randomly select 20 sessions in ACIDS dataset, and visualize their samples in a 2D space as shown in Figure~\ref{fig:before_and_after_sa-tcl}. In order to map the high dimensional features to 2D dots,  we simply concatenate and flatten the features from multiple modalities, and reduce the feature dimension to 50 by conducting principal component analysis (PCA), which accelerates t-SNE sufficiently without losing significant information that is essential for clustering.
Then we use t-distributed stochastic neighbor embedding (t-SNE) to further reduce the dimension to 2. We color the dots by their corresponding data collection session. After applying SA-TCL, samples from the same session are drawn closer together, while samples from different sessions are pushed farther apart. We also calculated their Silhouette Coefficient (marked as SC) as the metric for numerically measuring how "clustered" the samples are within the same session (higher values indicate higher clustering). This comparison demonstrates that SA-TCL mitigates the variances brought on by dynamic disturbances, thereby providing more distinguishable feature patterns to downstream classifiers.

\subsubsection{Frequency Mask}
In Table~\ref{tab:ablation_study_frequency_mask}, we remove the frequency mask in SA-TCL and evaluate the performance changes given DeepSense as the downstream classifier.
We observed that the frequency mask can further improve the performance brought by SA-TCL.
The frequency mask is empowered by our pre-knowledge regarding the information density across the frequency bands.
We speculate that it can provide a better parameter initialization for the neural network and help the temporal contrastive learning process focus on the high information density area of the spectrograms.
\input{content/tables/ablation_frequency_mask}

\subsection{Inference Cost of SudokuSens}
\begin{figure}
  \centering
  % \subfloat[FLOPs number]{
  %   \includegraphics[width=0.18\textwidth]{figure/evaluation/profiling/profiling_flops.pdf}
  %   \label{fig:profiling_flops}
  % }
  % \subfloat[Parameter number]{
  %   \includegraphics[width=0.20\textwidth]{figure/evaluation/profiling/profiling_param.pdf}
  %   \label{fig:profiling_params}
  % }
  % \subfloat[Time cost]{
  %   \includegraphics[width=0.20\textwidth]{figure/evaluation/profiling/profiling_time.pdf}
  %   \label{fig:profiling_time}
  % }
  \includegraphics[width=0.8\linewidth]{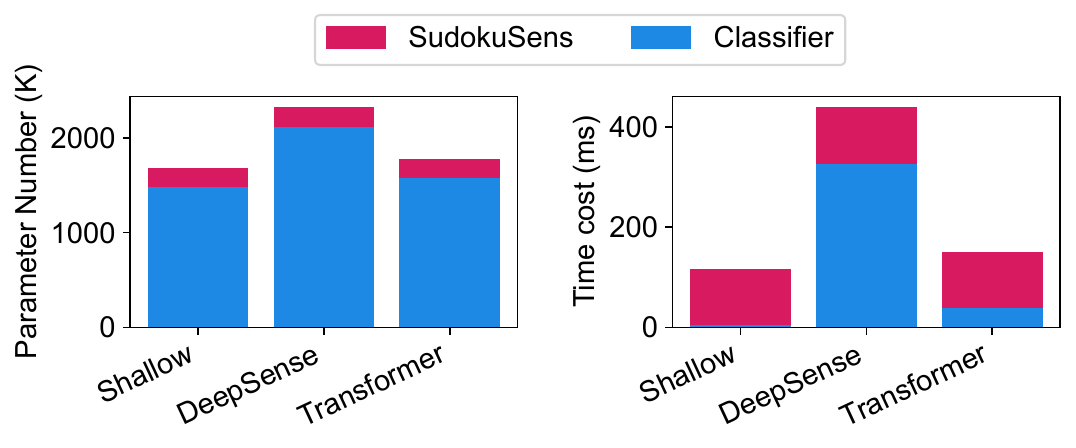}
  \vspace{-0.1in}
  \caption{Profiling of the execution cost of SudokuSens and the various downstream classifiers on Raspberry Pi 4.}
  \label{fig:profiling}
  \vspace{-0.1in}
\end{figure}

Only the encoder of the SA-TCL component in SudokuSens is utilized during the inference time, thereby allowing SudokuSens to act as a small-scale feature extractor before the downstream classifier in real-world IoT sensing deployment. We profile the number of parameters and the execution time cost on a Raspberry Pi 4, given input data collected during the case studies described in Section~\ref{subsec:case_study_performance}. The input batch size is set as 1.
Figure~\ref{fig:profiling} shows the results.

As shown in Figure~\ref{fig:profiling}, the number of parameters represents the memory space requirement of the neural network. SudokuSens, with its compact size compared to the three classifiers, can readily fit into the memory of a Raspberry Pi 4. Since the SA-TCL encoder in SudokuSens is primarily composed of convolutional layers, it generates higher floating point operations per second (FLOPS) than linear layers given the same number of parameters. As a result, SudokuSens has a higher execution time cost than both the shallow neural network and the Transformer, by 107 ms and 74 ms respectively, but it still runs 215 ms faster than DeepSense. DeepSense is notably slower than the other classifiers because of the overhead caused by inputting lengthy sequence data into its RNN layers. Considering the specific use case, where 2-second samples are used for vehicle type classification, any of the three classifiers can be integrated with SudokuSens for real-time vehicle detection. The small scale of SudokuSens during the inference stage enhances its practicality in a wide range of IoT applications.

%% file: content/tables/ablation_condinterp.tex
\renewcommand{\arraystretch}{1.3}
\begin{table}[h]
\caption{Influence of the number of interpolated samples to the performance of SudokuSens. The percentages after the dataset names represent the proportion of filled cells within the Sudoku matrix.}
\label{tab:ablation_study_condinterp}
\small
\resizebox{1.0\linewidth}{!}{
\fontsize{8.5}{11}\selectfont
\begin{tabular}{c|cc|ccccc}
\toprule
\multirow{2}{*}{\textbf{Dataset}} & \multicolumn{2}{l|}{\multirow{2}{*}{\textbf{\begin{tabular}[c]{@{}l@{}}Downstream\\ classifier\end{tabular}}}} & \multicolumn{5}{c}{\textbf{Interpolation ratio}} \\ \cline{4-8} 
 & \multicolumn{2}{l|}{} & \multicolumn{1}{c}{\textbf{0}} & \multicolumn{1}{c}{\textbf{0.1x}} & \multicolumn{1}{c}{\textbf{1x}} & \multicolumn{1}{c}{\textbf{2x}} & \multicolumn{1}{c}{\textbf{5x}} \\ \hline
\multirow{6}{*}{\textbf{\begin{tabular}[c]{@{}l@{}}ACIDS\\ (67\%)\end{tabular}}} & \multicolumn{1}{l|}{\multirow{2}{*}{\textbf{\begin{tabular}[c]{@{}l@{}}Shallow \\ Neural Network \end{tabular}}}} & \textbf{Acc} & 0.6113 & 0.6492 & 0.6544 & \textbf{0.6632} & 0.6280 \\
 & \multicolumn{1}{l|}{} & \textbf{F1} & 0.6002 & 0.6133 & \textbf{0.6588} & 0.6432 & 0.6363 \\ \cline{2-8} 
 & \multicolumn{1}{l|}{\multirow{2}{*}{\textbf{DeepSense}}} & \textbf{Acc} & 0.6299 & 0.6840 & \textbf{0.7251} & 0.7081 & 0.6486 \\
 & \multicolumn{1}{l|}{} & \textbf{F1} & 0.6039 & 0.6777 & \textbf{0.7252} & 0.7114 & 0.6300 \\ \cline{2-8} 
 & \multicolumn{1}{l|}{\multirow{2}{*}{\textbf{Transformer}}} & \textbf{Acc} & 0.5312 & 0.5820 & 0.6608 & \textbf{0.6884} & 0.6702 \\
 & \multicolumn{1}{l|}{} & \textbf{F1} & 0.5209 & 0.5957 & 0.6874 & \textbf{0.6933} & 0.6868 \\ \hline
\multirow{6}{*}{\textbf{\begin{tabular}[c]{@{}l@{}}Wearable-\\ HAR\\ (60\%)\end{tabular}}} & \multicolumn{1}{l|}{\multirow{2}{*}{\textbf{\begin{tabular}[c]{@{}l@{}}Shallow\\  Neural Network\end{tabular}}}} & \textbf{Acc} & 0.7021 & 0.7365 & \textbf{0.7490} & 0.7294 & 0.6541 \\
 & \multicolumn{1}{l|}{} & \textbf{F1} & 0.6411 & 0.6813 & \textbf{0.7688} & 0.7200 & 0.6540 \\ \cline{2-8} 
 & \multicolumn{1}{l|}{\multirow{2}{*}{\textbf{DeepSense}}} & \textbf{Acc} & 0.7158 & 0.8410 & 0.8613 & \textbf{0.8891} & 0.8363 \\
 & \multicolumn{1}{l|}{} & \textbf{F1} & 0.7055 & 0.8922 & 0.8888 & \textbf{0.9002} & 0.8516 \\ \cline{2-8} 
 & \multicolumn{1}{l|}{\multirow{2}{*}{\textbf{Transformer}}} & \textbf{Acc} & 0.6902 & 0.7158 & 0.7922 & 0.8194 & \textbf{0.8303} \\
 & \multicolumn{1}{l|}{} & \textbf{F1} & 0.7090 & 0.7014 & 0.8211 & 0.8001 & \textbf{0.8434} \\ \hline
\multirow{6}{*}{\textbf{\begin{tabular}[c]{@{}l@{}}Wi-Fi-HAR\\ (50\%)\end{tabular}}} & \multicolumn{1}{l|}{\multirow{2}{*}{\textbf{\begin{tabular}[c]{@{}l@{}}Shallow\\  Neural Network\end{tabular}}}} & \textbf{Acc} & 0.6427 & 0.7080 & \textbf{0.7492} & 0.7126 & 0.6688 \\
 & \multicolumn{1}{l|}{} & \textbf{F1} & 0.6512 & 0.6823 & \textbf{0.7337} & 0.6900 & 0.6448 \\ \cline{2-8} 
 & \multicolumn{1}{l|}{\multirow{2}{*}{\textbf{DeepSense}}} & \textbf{Acc} & 0.6676 & 0.7903 & \textbf{0.8616} & 0.8492 & 0.8111 \\
 & \multicolumn{1}{l|}{} & \textbf{F1} & 0.6811 & 0.7900 & 0.8711 & \textbf{0.8712} & 0.7919 \\ \cline{2-8} 
 & \multicolumn{1}{l|}{\multirow{2}{*}{\textbf{Transformer}}} & \textbf{Acc} & 0.6226 & 0.6814 & 0.754 & \textbf{0.7649} & 0.7402 \\
 & \multicolumn{1}{l|}{} & \textbf{F1} & 0.6220 & 0.6744 & 0.7105 & 0.7364 & \textbf{0.7550} \\ \hline

\end{tabular}}
\end{table}

%% file: content/tables/ablation_sa-tcl.tex
\renewcommand{\arraystretch}{1.3}
\begin{table}[h]
\caption{Performance improvement from SA-TCL. The percentages after the dataset names represent the proportion of filled cells within the Sudoku matrix.}
\label{tab:ablation_study_satcl}
\scriptsize
\resizebox{0.9\linewidth}{!}{
\fontsize{6}{8}\selectfont
\begin{tabular}{c|cc|cc}
\toprule
\textbf{Dataset} & \multicolumn{2}{c|}{\textbf{\begin{tabular}[c]{@{}c@{}}Downstream\\ classifier\end{tabular}}} & \textbf{w/o} & \textbf{w/} \\ \hline

\multirow{6}{*}{\textbf{\begin{tabular}[c]{@{}c@{}}ACIDS\\ (67\%)\end{tabular}}} & \multicolumn{1}{c|}{\multirow{2}{*}{\textbf{\begin{tabular}[c]{@{}c@{}}Shallow Neural \\ Network\end{tabular}}}} & 
\textbf{Acc} & 0.6052 & \textbf{0.6544} \\
 & \multicolumn{1}{c|}{} & \textbf{F1} & 0.6131 & \textbf{0.6588} \\ \cline{2-5} 
 & \multicolumn{1}{c|}{\multirow{2}{*}{\textbf{DeepSense}}} & \textbf{Acc} & 0.6773 & \textbf{0.7251} \\
 & \multicolumn{1}{c|}{} & \textbf{F1} & 0.6540 & \textbf{0.7252} \\ \cline{2-5} 
 & \multicolumn{1}{c|}{\multirow{2}{*}{\textbf{Transformer}}} & \textbf{Acc} & 0.6400 & \textbf{0.6608} \\
 & \multicolumn{1}{c|}{} & \textbf{F1} & 0.6810 & \textbf{0.6874} \\ \hline
 
\multirow{6}{*}{\textbf{\begin{tabular}[c]{@{}c@{}}Wearable-HAR\\ (60\%)\end{tabular}}} & \multicolumn{1}{c|}{\multirow{2}{*}{\textbf{\begin{tabular}[c]{@{}c@{}}Shallow Neural \\ Network\end{tabular}}}} & \textbf{Acc} & 0.7119 & \textbf{0.7490} \\
 & \multicolumn{1}{c|}{} & \textbf{F1} & 0.7008 & \textbf{0.7688} \\ \cline{2-5} 
 & \multicolumn{1}{c|}{\multirow{2}{*}{\textbf{DeepSense}}} & \textbf{Acc} & 0.8330 & \textbf{0.8613} \\
 & \multicolumn{1}{c|}{} & \textbf{F1} & 0.8514 & \textbf{0.8888} \\ \cline{2-5} 
 & \multicolumn{1}{c|}{\multirow{2}{*}{\textbf{Transformer}}} & \textbf{Acc} & 0.7714 & \textbf{0.7922} \\
 & \multicolumn{1}{c|}{} & \textbf{F1} & 0.7690 & \textbf{0.8211} \\ \hline
\multirow{6}{*}{\textbf{\begin{tabular}[c]{@{}c@{}}Wi-Fi-HAR\\ (50\%)\end{tabular}}} & \multicolumn{1}{c|}{\multirow{2}{*}{\textbf{\begin{tabular}[c]{@{}c@{}}Shallow Neural \\ Network\end{tabular}}}} & \textbf{Acc} & 0.7321 & \textbf{0.7492} \\
 & \multicolumn{1}{c|}{} & \textbf{F1} & 0.7045 & \textbf{0.7337} \\ \cline{2-5} 
 & \multicolumn{1}{c|}{\multirow{2}{*}{\textbf{DeepSense}}} & \textbf{Acc} & 0.8377 & \textbf{0.8616} \\
 & \multicolumn{1}{c|}{} & \textbf{F1} & 0.8519 & \textbf{0.8711} \\ \cline{2-5} 
 & \multicolumn{1}{c|}{\multirow{2}{*}{\textbf{Transformer}}} & \textbf{Acc} & 0.7222 & \textbf{0.754} \\
 & \multicolumn{1}{c|}{} & \textbf{F1} & 0.7336 & \textbf{0.7105} \\ \bottomrule
\end{tabular}}
\end{table}

%% file: content/tables/ablation_frequency_mask.tex
% \usepackage{multirow}

\begin{table}[th]
\fontsize{6}{8}\selectfont
\caption{Performance improvement from frequency masks.}
\vspace{-0.05in}
\label{tab:ablation_study_frequency_mask}
\begin{tabular}{l|llllll}
\hline
\multirow{3}{*}{} & \multicolumn{6}{c}{\textbf{Dataset}}                                                                                                                                                                      \\ \cline{2-7} 
                  & \multicolumn{2}{l|}{\textbf{ACIDS  (67\%)}}                                      & \multicolumn{2}{l|}{\textbf{Wearable-HAR (60\%)}}                               & \multicolumn{2}{l}{\textbf{Wi-Fi-HAR (50\%)}}              \\ \cline{2-7} 
                  & \multicolumn{1}{l|}{\textbf{w/o}} & \multicolumn{1}{l|}{\textbf{w/}}     & \multicolumn{1}{l|}{\textbf{w/o}} & \multicolumn{1}{l|}{\textbf{w/}}     & \multicolumn{1}{l|}{\textbf{w/o}} & \textbf{w/}     \\ \hline
\textbf{Accuracy} & \multicolumn{1}{l|}{0.6800}       & \multicolumn{1}{l|}{\textbf{0.7251}} & \multicolumn{1}{l|}{0.8523}       & \multicolumn{1}{l|}{\textbf{0.8613}} & \multicolumn{1}{l|}{0.8494}       & \textbf{0.8616} \\ \hline
\textbf{F1}       & \multicolumn{1}{l|}{0.6766}       & \multicolumn{1}{l|}{\textbf{0.7252}} & \multicolumn{1}{l|}{0.8662}       & \multicolumn{1}{l|}{\textbf{0.8888}} & \multicolumn{1}{l|}{0.8270}       & \textbf{0.8711} \\ \hline
\end{tabular}
\vspace{-0.15in}
\end{table}

%% file: content/related-work.tex
\section{Related Work}
\label{sec:related_work}
The paper falls in the category of {\em data augmentation\/}
solutions. Data augmentation is a common approach used to alleviate the data scarcity problem in IoT sensing applications~\cite{wen2020time}. Traditional time-series data augmentation carried out in the time domain includes scaling, jittering, rotating, permuting, noise injection, and others~\cite{um2017data, le2016data, cui2016multi, wen2019time, um2017data}. Since IoT signals may exhibit stronger patterns in the frequency domain~\cite{yao2019stfnets}, data augmentation techniques in the time-frequency domain were also proposed~\cite{zhang2020data, steven2018feature, park2019specaugment}. Recent studies have explored learning-based augmentation approaches, such as synthetic data generation using generative adversarial networks (GANs)~\cite{esteban2017real, yoon2019time, ratner2017learning}. There is also research in generating IoT sensor data based on data sources from other domain~\cite{kwon2020imutube, zhang2020deep}. However, these methods only address general data scarcity. They do not handle the specific incomplete conditional sampling problem we covered for IoT datasets.
    
Contrastive learning is commonly used as a self-supervised learning approach on unlabeled datasets for model pre-training~\cite{jaiswal2020survey}. Many of the prior techniques are based on SimCLR~\cite{chen2020simple}. It takes two augmented views of the same signal as a positive pair and maximizes their similarity, while minimizing their similarity with other samples in the batch~\cite{tang2020exploring, liu2021semi, khaertdinov2021contrastive}. More customized contrastive learning designs for time-series data are also proposed. For example, TF-C~\cite{zhang2022self} considers both time and frequency domain similarities as a measure distance during the contrastive loss calculation. TS2Vec~\cite{yue2022ts2vec} performs contrastive learning in a hierarchical way in both instance-wise and temporal dimensions, in order to extract more robust features. The multi-modality characteristic of IoT sensing data is leveraged to customize the contrastive loss~\cite{deldari2022cocoa, ouyang2022cosmo}. Temporal contrastive learning involves constructing positive pairs by the temporal relationships of the signals~\cite{dave2022tclr}. For example, TNC~\cite{tonekaboni2021unsupervised} constructs positive pairs for time series data based on temporal neighborhood. TCL~\cite{hyvarinen2016unsupervised} splits time-series data into segments, and predicts the segment ID during the contrastive learning as a way to extract underlying representation. Despite the fact that constrastive learning for IoT was vastly investigated, SudokuSens is the first to apply temporal contrastive learning by data collection sessions in IoT.

%% file: content/conclusion.tex
\section{Conclusion}
\label{sec:conclusion}
In this paper, we introduced SudokuSens, a novel framework that addresses data scarcity in IoT sensing applications. SudokuSens employs a Conditional Variational Autoencoder (CVAE) for conditional interpolation, enabling the synthetic generation of data for missing intrinsic attribute combinations, thereby enriching the diversity of the training dataset. Further, it incorporates Session-Aware Temporal Contrastive Learning (SA-TCL) to mitigate the variability introduced by dynamic disturbances, effectively enhancing the learned feature patterns for downstream classifiers. Across diverse IoT applications, SudokuSens has consistently demonstrated improved performance under unseen conditions, confirming its efficacy. This work advances robustness of intelligent IoT applications in real deployment scenarios.